\documentclass[10pt,twocolumn]{article}
\usepackage[letterpaper,margin=.75in,columnsep=.25in]{geometry}
\usepackage[T1]{fontenc}
\usepackage[utf8]{inputenc}
\usepackage{newtxtext}
\usepackage{microtype}
\usepackage[numbers,sort&compress]{natbib}

\usepackage{amsmath,amssymb,booktabs,array,tabularx}
\usepackage{graphicx,xcolor,enumitem,listings}
\usepackage{caption}
\usepackage[colorlinks=true,linkcolor=teal!65!black,citecolor=teal!65!black,urlcolor=teal!65!black]{hyperref}
\usepackage{placeins,float,needspace}
\setlist[itemize]{leftmargin=1.2em,itemsep=1pt,topsep=3pt}
\newsavebox{\papertablebox}
\newcommand{\fittable}[1]{\sbox{\papertablebox}{#1}\ifdim\wd\papertablebox>\linewidth\resizebox{\linewidth}{!}{\usebox{\papertablebox}}\else\usebox{\papertablebox}\fi}
\newcommand{\system}{\textsf{Synapse-Control}}
\newcommand{\authorentry}[3]{\parbox[t]{.30\textwidth}{\centering
{\fontsize{12}{14}\selectfont #1\par}
{\fontsize{9}{11}\selectfont #2\par #3\par}}}

\usepackage{stfloats}
\hypersetup{pdftitle={Synapse-Control: Making Agent Experience Actionable with Lightweight Runtime Control},
pdfauthor={Zhiyuan Huang, Shengan Zheng, Chenghu Yang, Song Zhang, Lingyun Li, Chen Yu, Weiquan Huang, Linpeng Huang},
pdfsubject={Agent experience and runtime control}}
\title{\vspace{-1.5em}\textbf{Synapse-Control: Making Agent Experience Actionable\\with Lightweight Runtime Control}}
\author{%
\begin{tabular}{@{}ccc@{}}
\authorentry{Zhiyuan Huang}{hzy364@sjtu.edu.cn}{Shanghai Jiao Tong University\\Shanghai, China\\Ant Group\\Hangzhou, China} &
\authorentry{Shengan Zheng\textsuperscript{*}}{shengan@sjtu.edu.cn}{MoE Key Lab of Artificial Intelligence,\\AI Institute, Shanghai Jiao Tong University\\Shanghai, China} &
\authorentry{Chenghu Yang}{yexiang.ych@antgroup.com}{Ant Group\\Hangzhou, China}
\\ \noalign{\vskip 10pt}
\authorentry{Song Zhang}{gepin.zs@antgroup.com}{Ant Group\\Hangzhou, China} &
\authorentry{Lingyun Li}{yunju.lly@antgroup.com}{Ant Group\\Hangzhou, China} &
\authorentry{Chen Yu}{yingjun.yc@antgroup.com}{Ant Group\\Hangzhou, China}
\\ \noalign{\vskip 10pt}
\authorentry{Weiquan Huang}{hwqqz678@sjtu.edu.cn}{Shanghai Jiao Tong University\\Shanghai, China\\Ant Group\\Hangzhou, China} &
\authorentry{Linpeng Huang\textsuperscript{*}}{lphuang@sjtu.edu.cn}{Shanghai Jiao Tong University\\Shanghai, China} &
\end{tabular}\\[5pt]
{\footnotesize\textsuperscript{*}Corresponding authors.}}
\date{}
\begin{document}
\maketitle
\begin{abstract}
Large language model agents reuse execution experience to improve reliability on complex tool-based tasks.
Systems such as Sentry call large models to generate failure diagnoses before retrieving advice, adding inference cost to experience reuse.
We present \system{}, an agent runtime whose lightweight control plane selects advice and determines when to send it to the task-executing model.
Rules shortlist relevant advice, and a lightweight 1.7B classifier fine-tuned on teacher-labeled state--suggestion pairs scores those candidates instead of generating a diagnosis.
Micro-contracts encode reusable interface checks, such as required arguments or valid options, without prescribing complete task programs.
A Prompt Value Table records earlier guidance and feedback to limit repeated suggestions through cooldown and token-budget checks.
Across 1,308 attempts per method in AppWorld and WebShop, \system{} achieves \textbf{270 successes (29.8\% more than Base)}, replaces \textbf{4,987 generative diagnosis calls}, and reduces auxiliary control time relative to Sentry by \textbf{50.2\% and 51.0\%}, respectively.
\end{abstract}
\section{Introduction}
\label{sec:introduction}

Reusing execution experience helps large language model agents perform complex tool-based tasks without repeatedly reconstructing procedures and interface requirements.
An agent alternates between reasoning, tool actions, and observations; its action-generating model, the \emph{actor}, supplies arguments, interprets returned state, and determines completion.
Existing systems retain skills, procedural insights, and lessons for subsequent tasks~\citep{voyager,expel,awm,sentry}.
Using this knowledge to correct a failed action or stalled execution is called \emph{runtime recovery}.
We focus on locally served agents that repeatedly use stable APIs or web interfaces, where both reliability and the cost of experience reuse matter.

Deciding when and how to reuse experience can itself require costly model inference.
Sentry examines the recent trajectory, generates a failure diagnosis, retrieves matching lessons, and verifies subsequent execution before publishing new experience~\citep{sentry}.
Whenever this diagnosis path is triggered, it generates intermediate text before the actor can use an existing lesson.
That work can be disproportionate to the correction needed, such as supplying a required argument.
Retrieved advice can also repeat a completed check or conflict with the current state.
Reducing this overhead requires both cheaper selection and a check that the advice is still useful.

Our central insight is a computational asymmetry between generating a diagnosis and judging a supplied suggestion.
Diagnosis generation constructs an explanation through successive token predictions; classification assigns scores to fixed labels for an already retrieved candidate.
\system{} exploits this asymmetry by using execution evidence and keyword matches to retrieve at most five suggestions, then learning only their suitability in the current state.
The 1.7B sidecar thus solves a different computational problem from the 9B detector: direct assessment replaces diagnosis generation.
Selection and delivery form the control plane, while the 9B actor retains planning and action generation.

Task-specific training makes this decomposition operational.
We fine-tune the complete 1.7B backbone and classification head on target-actor states collected with and without earlier guidance.
A 32B teacher labels the exact suggestion available at each state as useful, unnecessary, or conflicting.
The optimization objective concentrates adaptation on these recurring state--constraint distinctions.
The relevant requirement becomes discrimination at this decision boundary, rather than the breadth of a general response generator.

We represent reusable interface checks as \emph{micro-contracts}: short statements of local constraints and when they apply.
They specify facts such as required arguments or selected options while leaving task values and program composition to the actor.
Rules check for an unmet need and shortlist relevant entries; the classifier assesses the resulting suggestions.
Selected guidance is added to the actor's next-action prompt, which we call \emph{delivery}; the resulting actions and observations remain in its execution history.

We use a Prompt Value Table (PVT) to record each suggestion's delivery history and feedback within a task.
It withholds advice while feedback is pending, evidence is unchanged, or a cooldown is active.
Earlier feedback adjusts the ranking of eligible suggestions; token-budget checks and a single-hint-per-step limit constrain what reaches the actor.
These controls suppress repeated guidance when execution stalls.

We evaluate \system{} on AppWorld API programming and WebShop shopping tasks using the same actor and task workload for all primary methods.
Across 1,308 attempts per method, it achieves 270 successes, compared with 256 for Sentry and 208 for Base, with the lowest cumulative task time among these methods.
Auxiliary control time decreases by 50.2\% in AppWorld and 51.0\% in WebShop.
Controlled studies examine knowledge representation, candidate selection, and delivery history; the history ablation reduces repeated deliveries by approximately two thirds in both benchmarks.

Our contributions are:
\begin{itemize}
\item A decomposition of experience selection into rule-based retrieval and specialized candidate scoring, replacing 9B diagnosis generation with a fine-tuned 1.7B classifier.
\item A runtime combining micro-contracts, rule-based filtering, and delivery-history and budget management.
\item A two-benchmark evaluation of task success, benchmark score, and complete execution cost, supported by controlled component studies.
\end{itemize}

\section{Background and Motivation}
\label{sec:background}

Experience reuse extends an agent's execution loop with knowledge from earlier interactions; its efficiency depends on both knowledge construction and runtime use.
The main costs arise when the runtime diagnoses a failure, selects advice, and processes the resulting intervention.

\subsection{Experience and runtime recovery}

Execution experience allows an agent to reuse information about tool behavior and earlier outcomes across steps and tasks.
An interactive agent selects an action, executes it in the environment, and incorporates the returned observation into its next decision.
Actions can be API calls expressed as code or operations on a webpage, while observations reveal returned values, invalid arguments, unavailable actions, or a lack of progress.
ReAct and CodeAct illustrate this reasoning--action loop for tools and executable code~\citep{react,codeact}.
Retaining the procedures and checks learned from such interactions can reduce repeated exploration in later executions.

Runtime recovery uses corrective guidance to restore valid or productive execution after an error or a stall.
Stored interface checks, conditional instructions, and verified lessons supply this guidance.
For example, an API constraint can identify a missing argument, while a previous recovery lesson can suggest a check after repeated unsuccessful reads.
The agent still generates and executes the next action; the recovery mechanism supplies information for that decision.

Experience systems construct reusable knowledge in several forms.
Voyager stores executable skills, ExpeL extracts insights from trajectories, Agent Workflow Memory retains procedures, and ACE maintains an evolving playbook~\citep{voyager,expel,awm,ace}.
These representations range from complete programs to conditional instructions.
Their common runtime requirement is to connect stored knowledge to the current execution state before using it.

\begin{figure}[H]
\centering
\includegraphics[width=\linewidth]{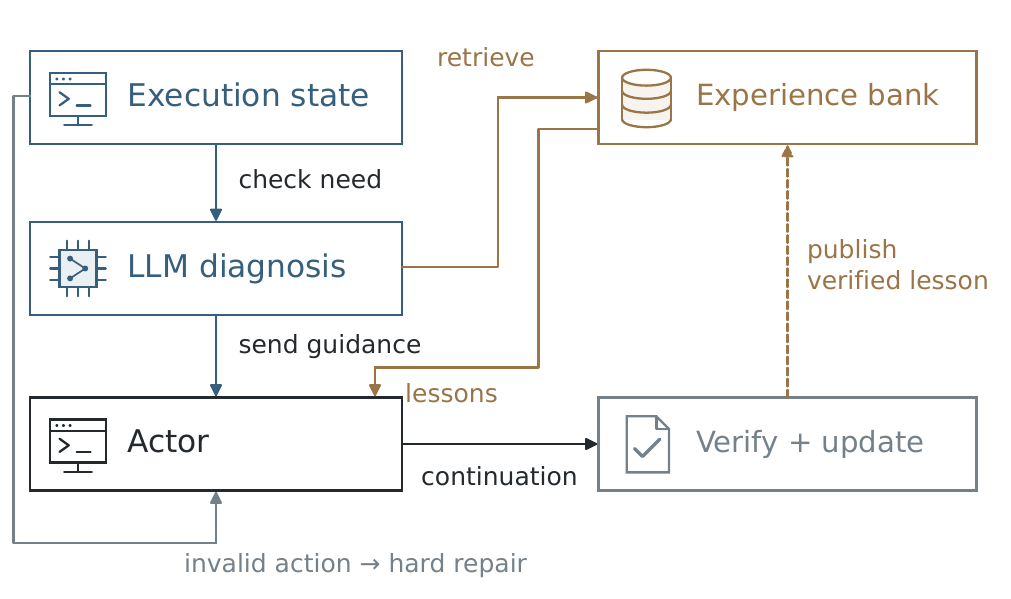}
\caption{Sentry's recovery workflow. Solid arrows show the guidance path; the dashed arrow publishes verified lessons.}
\label{fig:recovery-background}
\end{figure}

Sentry organizes recovery around a diagnosis--guidance--verification lifecycle~\citep{sentry}.
As shown in Figure~\ref{fig:recovery-background}, invalid actions enter a hard-repair path that handles execution errors directly; for other failures, a large-model monitor examines the recent trajectory and generates a diagnosis containing retrieval labels and local evidence.
The runtime then retrieves matching playbook entries and constructs guidance for the actor.
A verifier checks the subsequent execution, and verified recovery produces a new reusable lesson.
This design already includes intervention-need checking and an external experience store; the remaining question is the cost and precision of the decisions connecting them.

\subsection{The cost of diagnosis and verification}

Sentry makes auxiliary model calls to diagnose failures, verify corrections, and generate lessons.
The actor additionally processes the delivered guidance and may follow a different action sequence, so complete task time captures both control work and the resulting execution.
Separating these quantities identifies which overhead can be reduced by changing the control mechanism.

\begin{figure*}[!tbp]
\centering
\includegraphics[width=.94\linewidth]{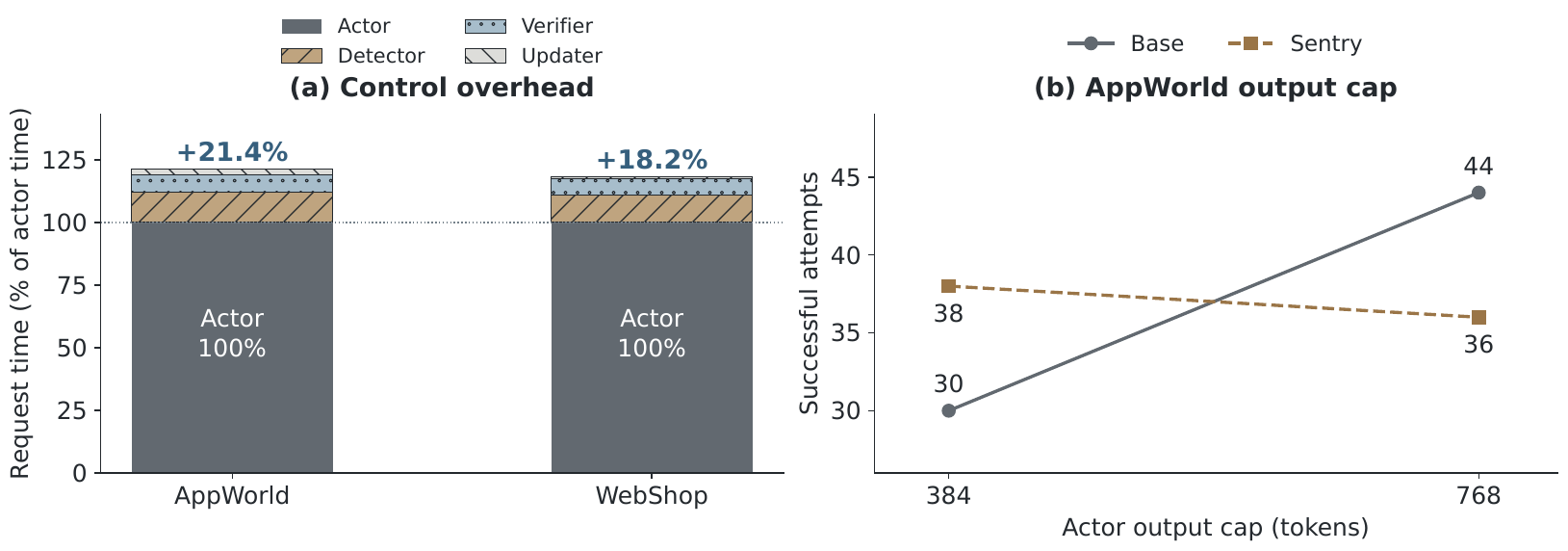}
\caption{\textbf{Recovery requires both economical control and effective guidance.} (a) Sentry's actor request time is normalized to 100\% within each benchmark; stacked segments show additional control work. (b) In the AppWorld development probe (342 attempts per condition), a larger actor-output cap increases Base completions but does not improve Sentry completions.}
\label{fig:motivation}
\end{figure*}

Sentry's measured control workload is substantial relative to actor inference in both environments (Figure~\ref{fig:motivation}a).
Auxiliary requests consume time equivalent to \textbf{21.4\% of actor request time in AppWorld and 18.2\% in WebShop}.
Actor request time averages 47.29 and 47.85 seconds per attempt, with 10.10 and 8.72 auxiliary seconds, respectively.
Generated diagnosis accounts for 5.74 and 5.21 auxiliary seconds; verification and lesson updates account for the remainder.
Normalizing to actor request time exposes the scale of repeated control work within each environment.

Repeated diagnosis is a concrete optimization target when relevant knowledge is already available.
A runtime with a small candidate pool needs to assess specific suggestions against visible evidence, yet generative diagnosis first produces an intermediate explanation and retrieval labels.
Assessing the candidate directly can reduce control work while preserving verification and lesson publication.

\subsection{State-dependent suitability of retrieved knowledge}

Topical relevance and present suitability answer different questions about a retrieved knowledge item.
\emph{Relevance} asks whether the item concerns the same tool, operation, or task topic; \emph{suitability} asks whether following its concrete advice addresses an unmet need without conflicting with the current state.
Lexical matching and interface identity support the former, whereas the latter requires the visible action, observed values, and checks already completed.

Visible execution state can change suitability even when the task topic remains unchanged.
In a shopping task, advice to change an option may match the product vocabulary but be unnecessary after the requested color has already been selected.
A generic instruction to search again can likewise redirect an otherwise ready purchase.
In API programming, an application-name match alone cannot distinguish a missing authentication fact from a pagination or argument constraint.
These cases require evaluating the proposed content against the current evidence, rather than treating retrieval as authorization to intervene.

Knowledge granularity and delivery history further determine how retrieved advice affects execution.
A required argument or visible-state check can be represented as a compact fact without prescribing an entire task program.
Such a fact still needs an applicability condition, and previously delivered advice may remain unresolved or redundant.
The runtime must therefore combine content assessment with a separate decision about repeated presentation.

Additional actor-generation capacity does not by itself ensure more effective recovery guidance.
In a separate AppWorld development probe (342 attempts per condition), increasing the actor-output cap from 384 to 768 tokens raises Base completions from \textbf{30 to 44}, while Sentry completions decrease from \textbf{38 to 36} (Figure~\ref{fig:motivation}b; Appendix~\ref{app:budget}).
This response motivates evaluating guidance at the state and candidate level rather than relying on a larger actor-output budget to resolve control errors.
The resulting design requirement is to turn stored knowledge into timely, actionable guidance without repeatedly generating an intermediate diagnosis.

\FloatBarrier

\section{The Synapse-Control Architecture}
\label{sec:system}

\begin{figure*}[b]
\centering
\includegraphics[width=\linewidth]{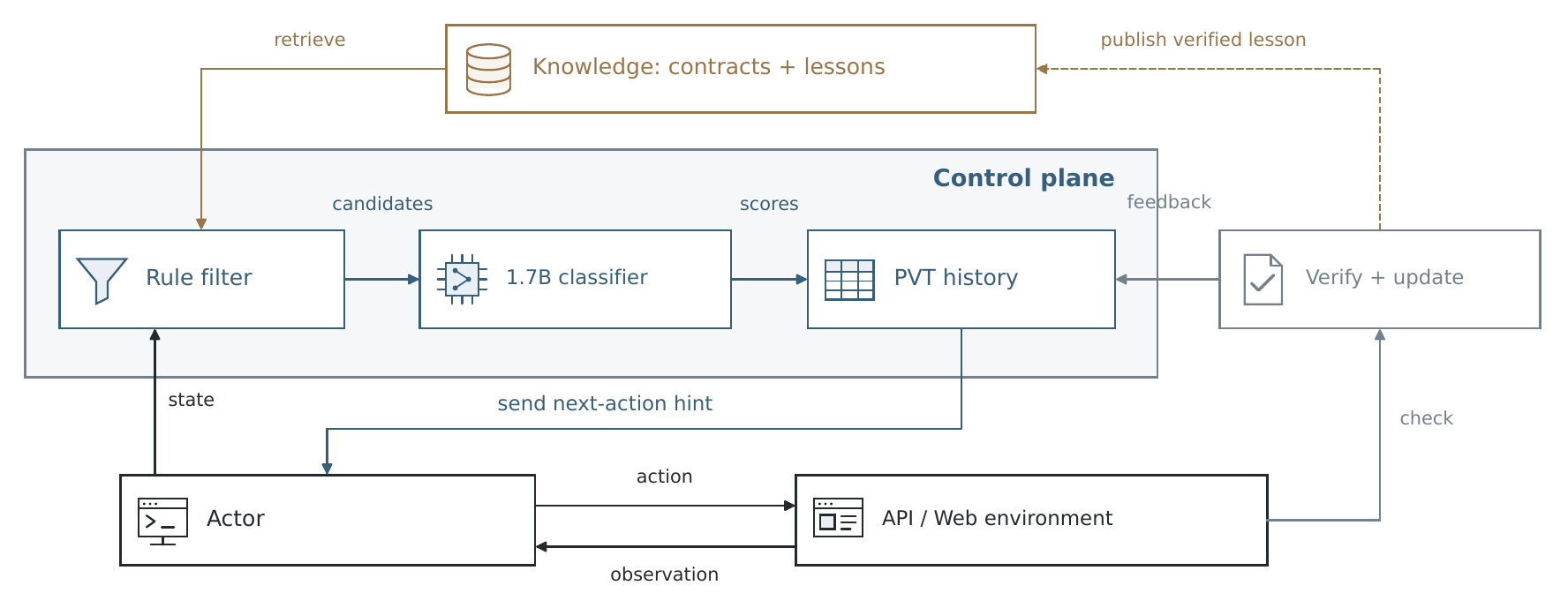}
\caption{\system{} overview. Rules filter and prepare advice (\S\ref{sec:assets}), the classifier scores it (\S\ref{sec:learning}), and PVT decides when to send it (\S\ref{sec:pvt}). Host verification and lesson updates remain active; hard repair is unchanged.}
\label{fig:architecture}
\end{figure*}

\system{} selects advice through a small classifier, represents interface checks as micro-contracts, and manages repeated use through PVT.
Figure~\ref{fig:architecture} connects these components to the actor and the existing recovery host, which performs verification and lesson updates.

The bank stores reusable entries, including micro-contracts and verified lessons.
Combining an entry with current evidence produces the \emph{candidate payload}: the exact suggestion the classifier evaluates.
If selected and permitted by PVT, this text enters the next-action prompt as a \emph{soft hint}, without changing the actor's action interface.

The execution plane generates task actions, while the control plane evaluates candidate payloads and governs soft-hint delivery. The classifier replaces the generative detector; host verification and lesson updates remain active. Let $s_t=\langle g,w_t,e_t\rangle$ denote the task objective, recent visible execution window, and evidence derived from that window. Deterministic rules extract this evidence from actions and observations: a rejected API call, unchanged reads, or a mismatch between requested and selected options. State construction requires no additional model-generated diagnosis.

\subsection{Learning to assess candidate guidance}
\label{sec:learning}

The architecture replaces a variable-length generative decision with fixed-output semantic classification.
For each candidate, a forward pass produces three label scores without autoregressively decoding a diagnosis.
At most five candidates and 1,024 tokens per input bound scoring work by configuration rather than knowledge-bank size.
All backbone parameters and the classification head are fine-tuned on the same state--suggestion objective, concentrating adaptation on distinctions needed by the target actor.
On held-out Contract candidates, the specialized scorer achieves 94.0\% Helpful admission precision in AppWorld and 98.15\% in WebShop; Section~\ref{sec:selector} reports coverage and downstream execution costs.

\paragraph{State--candidate classification.}
For state $s_t$ and candidate $c$ with its evidence-bound payload, the classifier estimates
\begin{equation}
p_\theta(y\mid s_t,c),\quad
y\in\{\mathrm{Helpful},\mathrm{Irrelevant},\mathrm{Harmful}\}.
\label{eq:classifier}
\end{equation}
Helpful denotes advice useful in the visible state; Irrelevant includes unnecessary advice; Harmful denotes conflicting or disruptive advice. A candidate passes the learned eligibility check when its calibrated Helpful probability exceeds the fixed operating threshold. Its priority is
\begin{equation}
v_t(c)=p_\theta(\mathrm{Helpful}\mid s_t,c)
       -p_\theta(\mathrm{Harmful}\mid s_t,c).
\label{eq:priority}
\end{equation}
An explicit Harmful class gives conflicting advice its own supervision and ranking penalty. Equation~\eqref{eq:priority} is the expected label utility under values $+1,0,-1$ for Helpful, Irrelevant, and Harmful: irrelevant advice adds no utility, while harmful advice reduces it. Ranking follows eligibility; identical difference scores do not admit two low-confidence candidates.

\begin{figure*}[b]
\centering
\includegraphics[width=\linewidth]{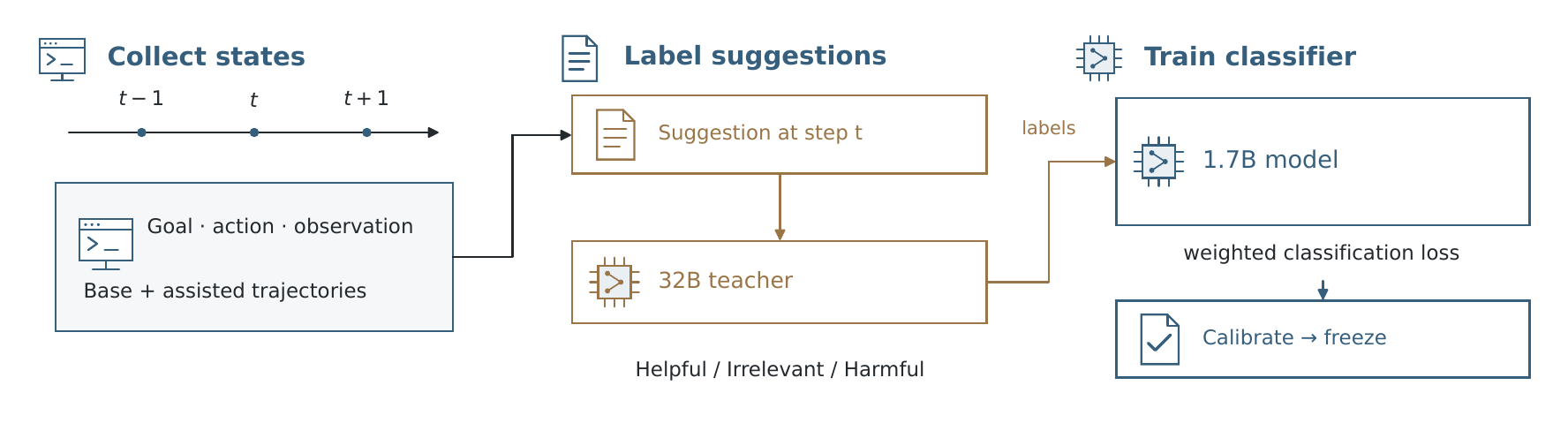}
\caption{Aligned supervision: collect target-actor states, reconstruct available candidate groups, label the exact payload, and fit the small classifier. Scenario or product-group partitions separate training, calibration, and classifier testing.}
\label{fig:training}
\end{figure*}

\paragraph{Sampling target-actor states.}
Training combines unassisted and recovery-assisted trajectories to cover states reached by the target actor under both policies.
All trajectories come from official training scenarios: Base contributes unassisted states, while recovery runs contribute states affected by earlier guidance.
This supervised policy mixture follows the distributional motivation for dataset aggregation~\citep{dagger}, since interventions change the states encountered later.

State sampling covers failure or stagnation, completion boundaries, and ordinary progress so that the classifier encounters both unmet needs and situations requiring no advice. These are state categories rather than assigned usefulness labels: a failure can lack suitable guidance, while a progressing task can make a suggestion redundant. Complete candidate groups are retained under a pair budget, preserving the alternatives at each sampled decision.

\paragraph{Reconstructing available suggestions.}
Reconstruction uses the online candidate builder and recorded publication order to reproduce the knowledge available at each decision. For initial bank $\mathcal K_0$ and lesson publication event $\tau(c)$, the available pool is
\begin{equation}
\mathcal K_t=\mathcal K_0\cup\{c:\tau(c)\prec t\}.
\label{eq:timevalid}
\end{equation}
Here $\tau(c)$ is the verified publication event of lesson $c$, $\prec$ is logical order within the replay stream, and $t$ denotes the current decision event. Only previously published lessons are available. The renderer $\mathcal R(s_t,c)$ binds current evidence to candidate content, producing the exact payload used for both teacher labeling and online delivery:
\begin{equation}
G_t=\{(s_t,\mathcal R(s_t,c),y_{t,c}):
                c\in\operatorname{Admit}(s_t,\mathcal K_t)\}.
\label{eq:training-group}
\end{equation}
The 32B teacher evaluates each state with the exact proposed suggestion, using visible values to distinguish useful checks from redundant or conflicting advice. Figure~\ref{fig:training} summarizes the training process.

The training objective weights the three usefulness classes to account for their different frequencies:
\begin{equation}
\mathcal L(\theta)=-\sum_{i=1}^{N_{\mathrm{train}}} w_{y_i}
       \log p_\theta(y_i\mid s_i,c_i),\qquad
w_y\propto n_y^{-1/2}.
\label{eq:loss}
\end{equation}
$N_{\mathrm{train}}$ is the training-pair count, and $n_y$ is the number of training labels of class $y$. Class weights use inverse square-root frequency. Calibration selects a temperature by negative log-likelihood on a held-out partition~\citep{calibration}; the deployment admission threshold is fixed at 0.90. The actor-state mixture and exact-payload labels focus learning on the judgments exercised by the runtime. Training and deployment configurations are specified in Section~\ref{sec:implementation} and Appendix~\ref{app:configuration}.

\subsection{Micro-contracts and candidate filtering}
\label{sec:assets}
\label{sec:gate}

Micro-contracts preserve the actor's planning autonomy by specifying \emph{what must hold} without prescribing \emph{how the task must be solved}.
Executable skills and reusable workflows encode an action structure~\citep{voyager,awm}; reusing such procedures requires reconciling their ordering and state assumptions with the current task.
Micro-contracts reduce this coupling by exposing local interface constraints that the actor can satisfy within its own plan.
Argument synthesis, variable bindings, and workflow composition remain with the actor.
The control plane supplies checks at the interface boundary instead of substituting an external program for task planning.
Rules match applicability conditions against execution evidence to shortlist suggestions.

\paragraph{Contract contents.}
An API micro-contract records the facts needed to check a local call:
\begin{equation}
\mu=\langle f,\mathcal K_{\mathrm{req}},
               \mathcal T_{\mathrm{args}},D_{\mathrm{return}}\rangle,
\label{eq:microcontract}
\end{equation}
where the fields identify the callable, required argument names, documented types and constraints, and documented return information. A candidate combines these facts with an applicability condition. WebShop uses the same interface for visible actions, selected options, and purchase readiness.

Knowledge assets have three sources: authored generic checks, structured interface documentation, and verified online lessons. In AppWorld, the builder reads parameters and return conventions from public API documentation and uses interfaces exercised in official training trajectories to construct reusable entries. At runtime, the builder binds visible task values to produce each candidate payload. In WebShop, generic assets cover option state, empty search, visible actions, evidence checks, and purchase readiness; the adapter supplies current page state.

\begin{figure*}[!tbp]
\centering
\includegraphics[width=\linewidth]{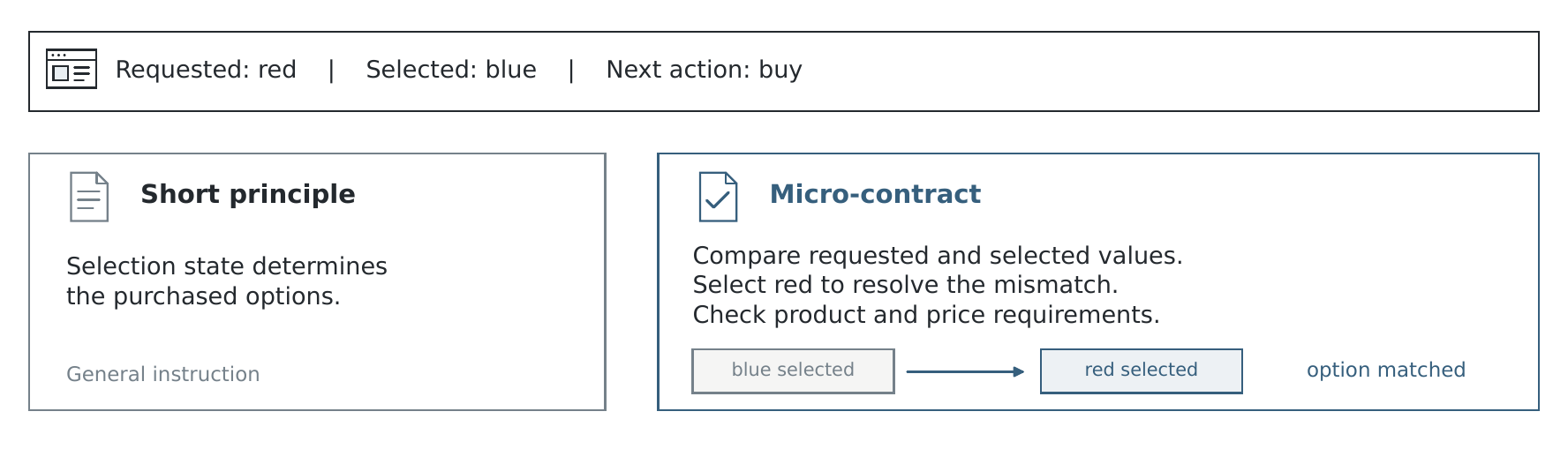}
\caption{Illustrative rendering of the implemented WebShop option-state asset. Both views use the same visible evidence; Contract adds the concrete check.}
\label{fig:contract}
\end{figure*}

Domain-level authoring defines generic checks and applicability conditions, while task-level binding is automated: WebShop uses five asset families, and AppWorld uses generic assets plus documentation mappings. \textbf{The authors estimate a few person-hours to define and organize one domain's templates using the existing adapters and builders.} The builder parses documented parameter and return fields, while the adapter binds visible options and local evidence. Delivery and lesson updates run automatically during evaluation, without task-specific rule authoring. Generic entries are maintained with the environment interface version; the actor still supplies task values and composes its program. Figure~\ref{fig:contract} contrasts a general principle with an actionable check.

\paragraph{Automatic contract extraction.}
The existing documentation builder provides a starting point for automating domain onboarding.
For an OpenAPI/Swagger interface, a static extractor can map the operation and parameter schemas to $f$, $\mathcal K_{\mathrm{req}}$, and $\mathcal T_{\mathrm{args}}$, and preserve documented response information in $D_{\mathrm{return}}$.
For a Python-facing API, the adapter must also map the documented operation to the exposed callable.
These fields can be copied from the versioned specification rather than authored for individual tasks.

An offline instruction-tuned 32B or 72B model could supplement that parser by converting natural-language descriptions into proposed applicability conditions and concise checks.
Each proposed condition would retain its source passage for review.
A validation pass would check the output schema, resolve endpoint and parameter references, and compare required fields, types, and enumerated values with the source specification.
Schema validation checks structure; a constraint inferred from prose also needs support in the cited documentation before admission.
Accepted skeletons would enter the same versioned bank and receive current values from the runtime adapter.
This is an extension of the construction workflow, with model inference confined to onboarding or interface updates; the experiments reported here use the documented builder and generic templates.

\paragraph{Rule filtering and learned scoring.}
The frontend uses execution evidence and lexical relevance to retrieve at most five candidates.
This inexpensive step supplies the alternatives; the specialized classifier only scores their concrete advice against the current state.
The candidate limit bounds scoring work independently of bank size.

A separate rule checks whether execution shows an unmet need before invoking the classifier:
\begin{equation}
\Gamma(s_t)=\mathbf1[
\mathrm{error}\lor\mathrm{repeat}\lor
\mathrm{stall}\lor\mathrm{gap}].
\label{eq:need}
\end{equation}
Under GPU colocation, every sidecar forward pass consumes device time and memory bandwidth shared with the actor.
Unconditional invocation therefore inserts additional GPU work between task actions, even for a small model.
$\Gamma(s_t)$ acts as a deterministic circuit breaker at the model-invocation boundary: once the trigger flags are extracted, an $O(1)$ Boolean decision admits or skips classification.
When the gate is closed, it issues zero classifier calls; local evidence checks remain active.
This makes neural control demand-driven and avoids unnecessary inference on the actor's execution path.

The terms refer to rejected actions, repeated action--observation pairs, unchanged documented reads, and visible information or commitment gaps. AppWorld checks three consecutive identical action--observation steps, or the same set of known GET interfaces with unchanged observations; unknown effects and writes do not use the GET relaxation. WebShop checks concrete option mismatches, empty results, unavailable actions, and pre-purchase gaps. Passing the gate enables scoring; candidate suitability and delivery eligibility are still checked. Appendix~\ref{app:configuration} specifies the runtime sequence.

\paragraph{Transient guidance and persistent execution state.}
Each soft hint is exposed for the next actor action, limiting repeated external text while preserving the resulting execution state.
Actions and observations enter normal task history, and AppWorld retains its Python session.
The actor can therefore reuse code and observations produced after the hint without repeatedly receiving the same reminder.
The knowledge bank and PVT retain the metadata needed to reconsider the asset after a later recurrence.

\subsection{Delivery governance with PVT}
\label{sec:pvt}

To prevent repeated candidate acceptance from producing uncontrolled repeated guidance, we introduce a task-local Prompt Value Table (PVT).
PVT regulates delivery using feedback, evidence changes, cooldown, and a shared budget, separating a knowledge asset's availability from its current delivery eligibility.

\paragraph{Delivery eligibility and state records.}
Each asset has a task-local record containing its lifecycle state, last delivery and feedback times, evidence key, recovery episode, pending feedback, and utility.
An available asset remains ineligible while feedback is unresolved, evidence is unchanged, cooldown is active, or its rendered payload exceeds the remaining budget.

Let $A_\theta(s_t,c)$ be frozen learned eligibility: event triggers require $p_\theta(\mathrm{Helpful}\mid s_t,c)>0.90$; periodic triggers additionally require $p_\theta(\mathrm{Harmful}\mid s_t,c)<0.05$ and visible evidence. Let $B_t$ be the remaining guidance budget and $\ell(c)$ the rendered actor-token cost. Then
\[
\begin{split}
\mathcal E_t=\{c\in\operatorname{Admit}(s_t,\mathcal K_t):{}&
A_\theta(s_t,c),\ \ell(c)\le B_t,\\
&\operatorname{PVTeligible}_t(c)\}.
\end{split}
\]
Here $\operatorname{PVTeligible}_t$ checks lifecycle, feedback, cooldown, and evidence/episode keys. The decision is
\begin{equation}
c_t=
\begin{cases}
\arg\max_{c\in\mathcal E_t}[v_t(c)+\lambda u_t(c)],
 &\Gamma(s_t)=1,\ \mathcal E_t\ne\emptyset,\\
\bot,&\text{otherwise}.
\end{cases}
\label{eq:pvt}
\end{equation}
Eligibility requires $\ell(c)\leq B_t$. Delivery subtracts this cost and marks feedback pending; abstention leaves the delivery state unchanged. A new evidence/episode pair can permit reuse after cooldown. Additional hard references consume the same task-level budget.

\paragraph{Feedback-driven utility updates.}
PVT updates historical utility from the host verifier's assessment of local recovery, using execution evidence rather than benchmark rewards. Let $q_t(c)$ indicate a defined, enabled utility observation. Observed progress gives $r_t=+1$; regression or adoption without progress gives $r_t=-1$:
\begin{equation}
u_{t+1}(c)=
\begin{cases}
(1-\alpha)u_t(c)+\alpha r_t(c),&q_t(c)=1,\\
u_t(c),&q_t(c)=0.
\end{cases}
\label{eq:update}
\end{equation}
Unknown outcomes, ignored advice, and missing adoption evidence leave utility unchanged. Feedback can still advance lifecycle metadata and end a pending episode, separating delivery-state progress from the availability of a utility signal.

\paragraph{Budget and utility invariants.}
The controller provides three directly checkable properties, proved in Appendix~\ref{app:pvt-properties}. Let $\ell(c_t)$ be the token cost of delivered soft guidance, with $\ell(\bot)=0$, and let $h_t\geq0$ be the token cost of any hard-repair reference added at step $t$. First,
\begin{equation}
\sum_{t=1}^{T}[\ell(c_t)+h_t]\leq B_0.
\label{eq:budget-safety}
\end{equation}
Second, unresolved feedback leaves utility unchanged. Third, starting from $u_0=0$, updates preserve $u_t(c)\in[-1,1]$. Thus $|\lambda u_t(c)|\leq\lambda$. At the evaluated $\lambda=0.1$, each adjustment is at most 0.1, and the perturbation of any pairwise priority gap is at most 0.2. History reorders the high-confidence pool; suitability, cooldown, and budget eligibility are checked first, so historical success cannot bypass them.

\FloatBarrier

\section{Implementation}
\label{sec:implementation}

Each worker colocates the actor and classifier, constructs bounded state inputs, and routes verification and lesson updates through the existing recovery host.

\paragraph{Serving and state construction.}
Each evaluation worker colocates the Qwen3.5-9B actor service and Qwen3-1.7B classifier on one H20 GPU. Tensor parallelism is one; the actor service admits one active sequence and starts with memory utilization 0.65. The implementation replaces a generative request on the control path; it does not require another GPU for each actor.

Bounded state construction preserves the information needed for a state--candidate judgment within a fixed classifier input. The controller retains at most five recent execution steps, including bounded observation values and entity samples. It preserves the objective, current action, and complete candidate payload. Older steps are removed first, followed by bounded observation compaction. If the required state and payload do not fit, the candidate is skipped. Four limits govern different resources: 1,024 tokens per classifier input; 2,048 added guidance tokens per task; 768/384 actor-output tokens in AppWorld/WebShop; and a 40,960-token actor-service context.

\paragraph{Learning and delivery settings.}
Domain-specific fine-tuning converts Qwen3-32B teacher judgments into a frozen 1.7B suitability classifier. We train the full backbone and classification head with AdamW, BF16 parameters, and FP32 loss logits. The two domain classifiers are trained independently from the same initial checkpoint, using 456 optimizer updates for AppWorld and 450 for WebShop. The reset three-class checkpoints are selected on development data and frozen before final evaluation. Calibrated Helpful probability must exceed 0.90. PVT uses $\alpha=0.5$, $\lambda=0.1$, and a three-step cooldown. Deterministic item identifiers break ranking ties. Training splits, hyperparameters, teacher prompts, and realized sampling distributions appear in Appendix~\ref{app:configuration}.

\paragraph{Offline construction and online reuse.}
Offline preparation produces a domain controller reused across tasks without additional labeling or fine-tuning.
Artifact filesystem timestamps give an approximately 13-minute AppWorld training
batch: two heads run concurrently on two H20 GPUs, each with 456 optimizer updates,
with loading, evaluation, and checkpoint persistence included. WebShop's 450-update
domain model takes approximately 15 minutes for training, evaluation, and persistence.

WebShop's 3,000-pair labeling stage takes approximately 38.5 minutes. Data assembly
produces 2,998 labeled examples: 2,398 for training and 300 each for calibration and
testing. The complete chain from labeling launch to checkpoint persistence takes
approximately 54 minutes. Appendix~\ref{app:configuration} specifies the stage
boundaries and timing scope.

The resulting checkpoints remain frozen across 1,008 AppWorld and 300 WebShop
evaluation attempts. Relative to Sentry, these workloads reduce cumulative task
execution time by approximately 53.5 and 65.4 minutes, respectively. Online use
reuses the classifier; the host verification and publication path updates experience
without relabeling or retraining for each new task. A single domain preparation therefore supports the full task workload, spreading its one-time labeling and fine-tuning work across repeated use.

\paragraph{Recovery-host integration.}
Host integration preserves Sentry's hard repair, verification, and lesson updates while replacing its generative selection path. AppWorld preserves the task's Python session and uses the host's ten-step soft-verification horizon. Invalid execution can receive hard repair while a soft episode is pending. Newly attached hard references and soft guidance share one budget. The verifier sees the continuation and local recovery evidence, not benchmark rewards.

The WebShop adapter supplies the same visible action constraints to every evaluated method using text observations. Search requires a visible search input; clicks require a visible target. The search check supplements the original text interface. A rejected action returns feedback and consumes an actor step. Generic contract families cover option state, empty search, visible actions, evidence checks, and purchase readiness. Before a proposed purchase, visible unmet requirements can trigger the same scoring and PVT path. At termination, pending recovery can be checked against the available shortened continuation. All compared methods use the same adapter, actor budget, and terminal evaluator.

\section{Evaluation}
\label{sec:evaluation}

The evaluation tests whether lightweight experience control improves task success and execution efficiency, and then isolates the effects of knowledge representation, candidate selection, and delivery history.

\begin{table*}[t]
\centering\small
\fittable{\begin{tabular}{lrrrrrr}
\toprule
Method & \shortstack{Gen. diagnosis\\calls $\downarrow$} & \shortstack{Total successes\\$\uparrow$} & \shortstack{Success gain\\vs. Base $\uparrow$} & \shortstack{AppWorld\\Successes $\uparrow$} & \shortstack{WebShop\\Successes $\uparrow$} & \shortstack{Total task\\seconds $\downarrow$} \\
\midrule
Base & \textbf{0} & 208/1,308 & --- & 136 & 72 & 76,186.2 \\
Sentry & 4,987 & 256/1,308 & +23.1\% & 173 & 83 & 82,002.7 \\
\textbf{Synapse-Control} & \textbf{0} & \textbf{270/1,308} & \textbf{+29.8\%} & \textbf{178} & \textbf{92} & \textbf{74,874.0} \\
\bottomrule
\end{tabular}
}
\caption{Primary results over 1,308 attempts per method (1,008 AppWorld; 300 WebShop). Successes satisfy all task requirements; diagnosis counts generative detector calls. Task time includes all actor and control work. Benchmark scores and paired gains appear in Section~\ref{sec:quality}.}
\label{tab:main}
\end{table*}

\subsection{Methodology}
\label{sec:protocol}

\paragraph{Benchmarks and methods.}
AppWorld evaluates stateful API programming~\citep{appworld}. We use all 168 \texttt{test\_normal} tasks across 56 scenarios and six task orders: 1,008 attempts per method. WebShop evaluates search, inspection, option selection, and purchase~\citep{webshop}. Its main reevaluation contains 100 tasks and three task orders: 300 attempts per method.

The primary methods are \textbf{Base}, the unassisted Qwen3.5-9B actor; \textbf{Sentry}, which generates diagnoses to retrieve advice; and \textbf{\system{}}, our learned selection and delivery runtime. Sentry denotes our integration of the published recovery core with the common benchmark harness. Each comparison fixes the actor, task order, action interface, and terminal evaluator. This controls actor capability while comparing recovery policies. Further knowledge-matched controls are defined and evaluated in Appendix~\ref{app:ablation}.

Each repeat begins with its declared initial bank; verified lessons accumulate within the repeat. The three primary methods register 3,924 attempts, and the additional knowledge-matched condition brings the total to 5,232. Actor temperature is 0.7, with at most 60 actions per task. AppWorld uses a 768-token actor-output cap; WebShop uses 384 tokens and minimal zero-shot text prompts. AppWorld training and checkpoint selection precede held-out evaluation. WebShop reevaluates a fixed 100-task prefix after development and validation. Appendix~\ref{app:configuration} specifies both protocols.

\paragraph{Measurements.}
Complete success means all AppWorld task checks pass, or WebShop reward equals one. Score is mean test-pass percentage or reward times 100. Means use scored attempts; completion counts and costs retain every registered attempt. Auxiliary control time sums detector, verifier, updater, and sidecar elapsed time. End-to-end task time measures the complete execution. Request timings include client and service waiting; local inference is timed at the worker. Amortized cost per completion divides total registered-workload cost by complete successes.

\subsection{Main quality and cost results}
\label{sec:quality}\label{sec:cost}

\paragraph{Success yield and execution efficiency.}
\system{} records \textbf{270 successes}, versus 256 for Sentry and 208 for Base: \textbf{29.8\% more than Base} on the fixed workload (Table~\ref{tab:main}). It also records the lowest cumulative task time. AppWorld contributes \textbf{178 successes} versus 173 for Sentry and 136 for Base; WebShop contributes \textbf{92} versus 83 and 72, increasing successes over Sentry by \textbf{5 and 9}, respectively.

Mean scores for Base, Sentry, and \system{} are 50.32, 56.23, and 55.95 in AppWorld, and 48.23, 56.04, and 55.82 in WebShop, respectively. Thus, \system{} achieves more successes with mean scores \textbf{within 0.3 points of Sentry in both benchmarks}. On jointly scored task--repeat pairs, score gains against Base are +5.69 and +7.60; every repeated-order comparison is positive.

\begin{figure*}[b]
\centering
\includegraphics[width=.96\linewidth]{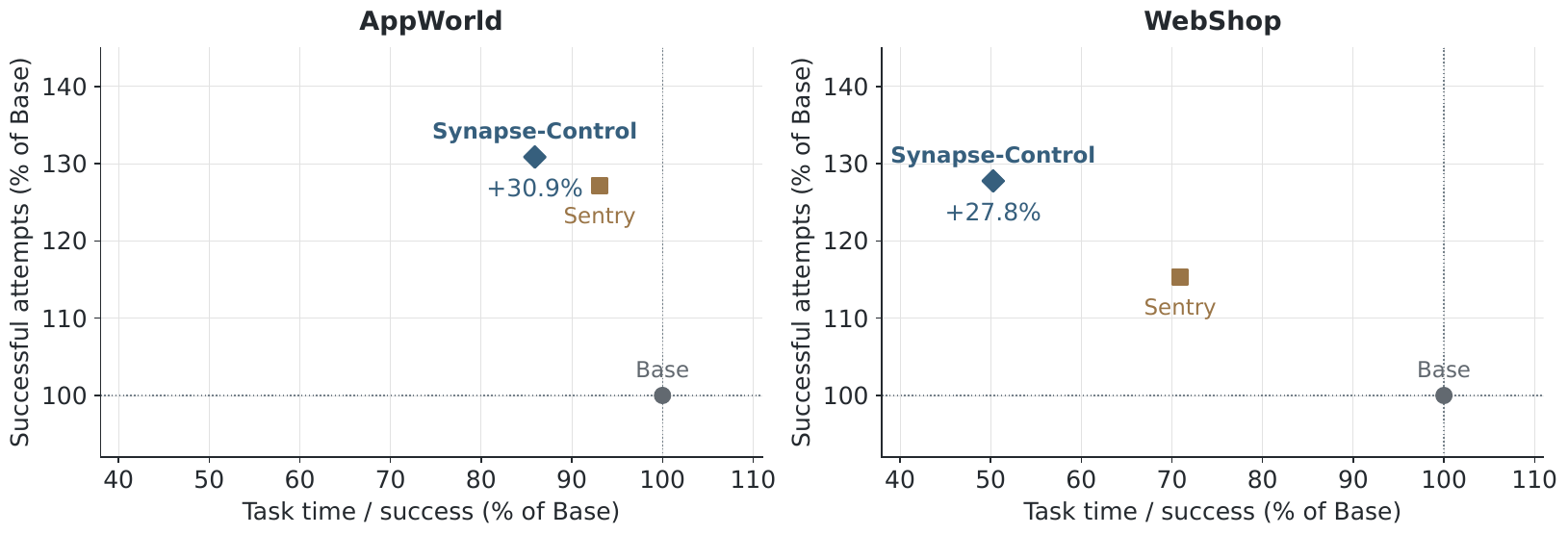}
\caption{\textbf{More completions with lower amortized execution cost.} Each benchmark normalizes completions and total task time per completion to Base (100\%). The cost numerator includes all registered attempts. Both panels use the same methods and axes.}
\label{fig:quality-cost}
\end{figure*}

Figure~\ref{fig:quality-cost} relates completion yield to complete execution cost. Relative to Base, total task time per completion falls by \textbf{14.1\% in AppWorld and 49.7\% in WebShop}, while completed attempts increase in both.

\begin{figure*}[t]
\centering
\includegraphics[width=.96\linewidth]{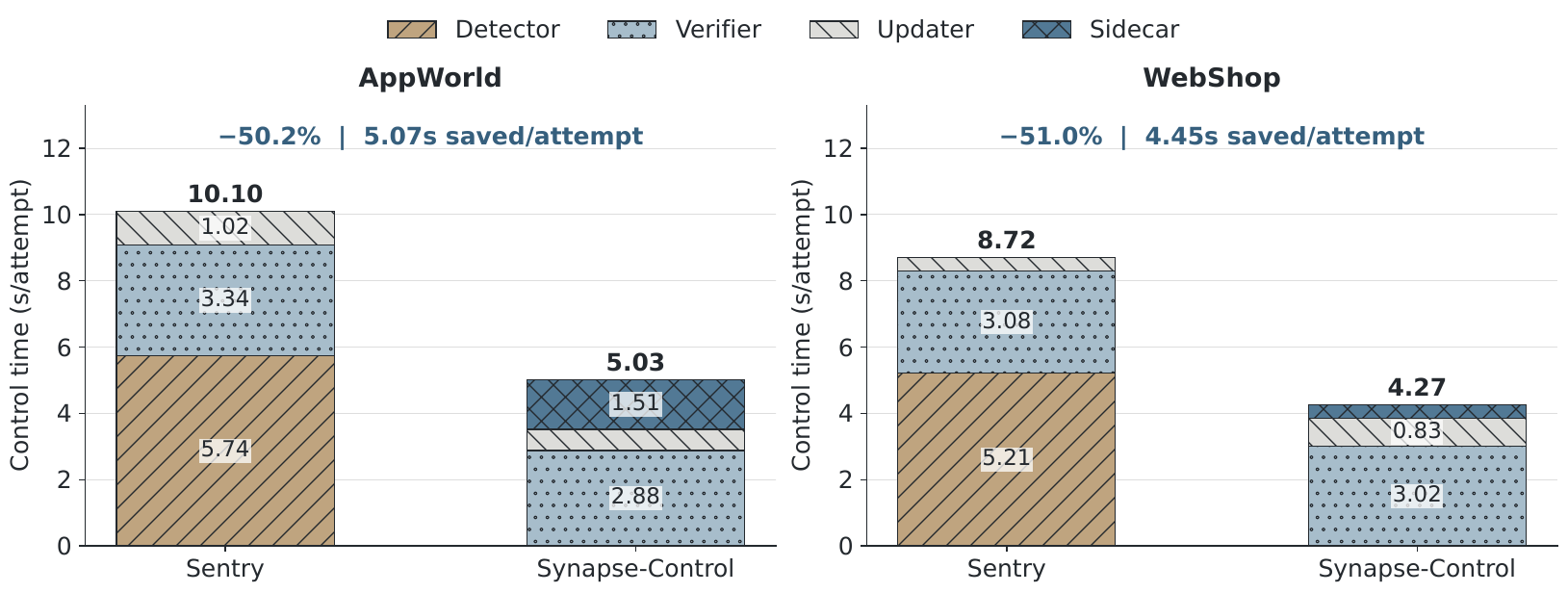}
\caption{\textbf{Control time is halved in both benchmarks.} Replacing generative diagnosis with the 1.7B sidecar reduces auxiliary control time by 50.2\% in AppWorld and 51.0\% in WebShop. Both panels show the same four roles per registered attempt; verification and lesson publication remain active.}
\label{fig:cost}
\end{figure*}

\paragraph{Control time is halved in both environments.}
The runtime replaces \textbf{4,987 generative detector calls}: 4,002 in AppWorld and 985 in WebShop. AppWorld auxiliary control time falls from \textbf{10.10 to 5.03 seconds per attempt (50.2\%)}, and WebShop falls from \textbf{8.72 to 4.27 seconds (51.0\%)}. Figure~\ref{fig:cost} uses the same four components, denominator convention, and vertical scale in both panels.

AppWorld auxiliary large-model requests fall from \textbf{6,536 to 2,048 (68.7\%)}. Its complete task time falls by \textbf{5.0\%} relative to Sentry. WebShop complete task time falls by \textbf{21.4\%} relative to Sentry and \textbf{35.8\%} relative to Base.

\begin{table*}[t]
\centering\small
\setlength{\tabcolsep}{4pt}
\fittable{\begin{tabular}{lrrrrrr}
\toprule
 & \multicolumn{3}{c}{AppWorld: 1,008 attempts/method} & \multicolumn{3}{c}{WebShop: 300 attempts/method} \\ \cmidrule(lr){2-4}\cmidrule(lr){5-7} Metric & Base & Sentry & \shortstack{Synapse-\\Control} & Base & Sentry & \shortstack{Synapse-\\Control} \\
\midrule
Complete successes & 136/1,008 & 173/1,008 & \textbf{178/1,008} & 72/300 & 83/300 & \textbf{92/300} \\
Large-model requests & 30,293 & 33,342 & 30,828 & 7,225 & 6,977 & \textbf{5,885} \\
Generative detector calls & 0 & 4,002 & \textbf{0} & 0 & 985 & \textbf{0} \\
Actor input tokens (M) & 234.92 & 226.59 & 246.66 & 105.67 & 71.63 & \textbf{62.34} \\
Auxiliary input tokens (M) & 0.00 & 21.90 & 8.09 & 0.00 & 6.82 & \textbf{3.47} \\
All service input tokens (M) & 234.92 & 248.49 & 254.75 & 105.67 & 78.45 & \textbf{65.81} \\
Input tokens / completion (M) & 1.727 & 1.436 & \textbf{1.431} & 1.468 & 0.945 & \textbf{0.715} \\
Task seconds / completion & 395.45 & 368.14 & \textbf{339.78} & 311.17 & 220.66 & \textbf{156.45} \\
Auxiliary control time (s) & 0.0 & 10,178.5 & \textbf{5,067.3} & 0.0 & 2,614.8 & \textbf{1,280.8} \\
Complete task time (s) & 53,781.7 & 63,687.6 & 60,480.3 & 22,404.5 & 18,315.2 & \textbf{14,393.6} \\
\bottomrule
\end{tabular}
}
\caption{Completed work and corresponding costs in both benchmarks. Per-completion metrics divide the entire registered workload's input or task time by complete successes. Timing follows Section~\ref{sec:protocol}.}
\label{tab:cost}
\end{table*}

\subsection{Controlled design comparisons}
\label{sec:ablation}

Controlled comparisons examine three design responsibilities: the content of guidance, the cost of selecting it, and the regulation of repeated delivery. Two AppWorld development probes isolate the knowledge interface and need gate (Appendix~\ref{app:ablation}); matched-knowledge and history comparisons report both benchmarks.

\paragraph{Actionable facts improve the guidance interface.}
Holding the controller and need gate fixed, the 318-attempt Short/Contract comparison changes the delivered content. Adding local interface facts to concise principles increases complete successes from \textbf{31 to 46 (+48.4\%)} and mean score from 50.32 to 53.05. This supports the micro-contract interface: the actor benefits from concrete constraints while retaining responsibility for its program.

\paragraph{Cost reduction persists under the same need gate.}
We give a knowledge-matched generative control the same observable necessity gate, with 342 registered attempts per arm. \system{} reduces auxiliary large-model requests from \textbf{1,436 to 572 (60.2\%)} and task time from 17,681.7 to 16,496.8 seconds (\textbf{6.7\%}). Aligning admission therefore preserves the efficiency advantage of the subsequent selection and delivery path. The gate alone does not account for the reduction.

\paragraph{Delivery bounds complement candidate selection.}
PVT constrains the delivery sequence: soft hints and added hard references share a 2,048-token task budget, at most one soft hint is exposed per step, and recurrence requires resolved feedback, new evidence, and a three-step cooldown. The operational arguments in Appendix~\ref{app:pvt-properties} establish these budget and recurrence guarantees independently of the classifier's ranking.

\paragraph{Control efficiency with matched knowledge.}
With the knowledge interface fixed, \system{} reduces auxiliary time by \textbf{50.2\% in AppWorld and 60.4\% in WebShop}. Complete success rates are 17.7\% versus the generative control's 16.2\% in AppWorld, and 30.7\% versus 31.7\% in WebShop; full task time falls by 8.4\% and 18.9\%, respectively. This comparison holds knowledge fixed while changing selection and delivery. Table~\ref{tab:shared} (Appendix~\ref{app:ablation}) defines the configuration and reports both benchmarks.

\paragraph{Operational delivery checks.}
The main logs contain checkable budget records for 993 AppWorld and 219 WebShop attempts, with no delivery exceeding the 2,048-token added-content budget. Soft hints and added hard references use the same accounting path, consistent with Equation~\eqref{eq:budget-safety}. Appendix~\ref{app:pvt-properties} summarizes the corresponding delivery and budget records.

\begin{table*}[tbp]
\centering\small
\fittable{\begin{tabular}{lrrrr}
\toprule
 & \multicolumn{2}{c}{AppWorld} & \multicolumn{2}{c}{WebShop}\\
\cmidrule(lr){2-3}\cmidrule(lr){4-5}
Metric & No-history & Full & No-history & Full\\
\midrule
Repeated guidance deliveries & 65 & \textbf{23} & 6 & \textbf{2}\\
Fewer repeat deliveries & --- & \textbf{64.6\%} & --- & \textbf{66.7\%}\\
Complete successes & 15/90 & \textbf{17/90} & \textbf{25/90} & \textbf{25/90}\\
Large-model requests & 2,859 & 2,900 & 1,890 & 1,699\\
Input tokens (M) & 23.71 & 24.82 & 22.20 & 19.52\\
Output tokens (M) & 0.36 & 0.38 & 0.32 & 0.28\\
\bottomrule
\end{tabular}
}
\caption{\textbf{History governance removes about two thirds of repeated deliveries in both benchmarks.} Each benchmark has 90 attempts per arm, all scored and paired. Costs retain all attempts. No-history preserves the budget and single-hint limit.}
\label{tab:pvt-ablation}
\end{table*}

\paragraph{History governance reduces repeated delivery.}
We compare full PVT (Full) with No-history on 90 matched task pairs per benchmark: 360 attempts, all scored. Six isolated workers each handle 15 distinct tasks per benchmark; both arms use the same tasks and order, with separate online experience. No-history disables utility ranking, the three-step cooldown, pending-feedback checks, and same-evidence/episode checks. It retains the classifier, contracts, need gate, 2,048-token budget, single-hint-per-step limit, and host verification and update path.

\textbf{History governance reduces repeated guidance deliveries from 65 to 23 in AppWorld (64.6\%) and from 6 to 2 in WebShop (66.7\%).}
AppWorld completions increase from 15 to 17. In WebShop, both arms complete 25 attempts, while Full uses \textbf{10.1\% fewer large-model requests and 12.1\% fewer input tokens}. Table~\ref{tab:pvt-ablation} reports corresponding success counts, repeated deliveries, and workload measurements for both domains.

Reason-level counts attribute 39 and 20 AppWorld candidate blocks to cooldown and same-evidence checks, respectively; WebShop records 52 and 6. Budget checks remain active in both arms and are counted separately. On the same reached states, a shadow scan at the deployed $\lambda=0.1$ changes the top-ranked candidate in 17 AppWorld decisions and none in WebShop. History participates in ranking in the former; its observed role in the latter lies in delivery eligibility. Appendix~\ref{app:pvt-ablation} reports the reason breakdown and three-weight ranking scan.

\subsection{Downstream efficiency of learned selection}
\label{sec:selector}

We evaluate learned selection at two levels: admission judgments on held-out candidate pairs and downstream quality and cost in paired agent executions.
The first measures agreement with teacher usefulness labels; the second measures the consequences of using the selected guidance.

\paragraph{Candidate admission precision.}
On classifier tests separated by scenario or product group from training, the frozen 0.90 operating point achieves \textbf{94.0\% Helpful admission precision in AppWorld and 98.15\% in WebShop} on Contract candidates, accepting 65.8\% and 72.0\% of candidates (Table~\ref{tab:classifier-contract}).
Precision measures agreement between admitted guidance and the teacher's usefulness labels on the deployed micro-contract input.
One harmful candidate is accepted among 76 AppWorld Contract candidates, and none among 150 WebShop candidates; Appendix~\ref{app:classifier-audit} provides content-stratified counts.

\begin{table}[tbp]
\centering\small\setlength{\tabcolsep}{3pt}
\fittable{\begin{tabular}{lrr}
\toprule
Metric & AppWorld & WebShop \\
\midrule
Test candidates & 76 & 150 \\
Accepted candidates & 50 & 108 \\
Helpful precision & \textbf{47/50 (94.0\%)} & \textbf{106/108 (98.15\%)} \\
Helpful recall & 47/65 (72.31\%) & 106/121 (87.60\%) \\
Coverage & 50/76 (65.79\%) & 108/150 (72.00\%) \\
\bottomrule
\end{tabular}
}
\caption{\textbf{Admission precision and coverage on deployed content.} Both benchmarks use Contract test candidates and frozen admission rules. Precision divides accepted Helpful candidates by all accepted candidates; recall divides accepted Helpful candidates by all Helpful candidates.}
\label{tab:classifier-contract}
\end{table}

Calibration sensitivity characterizes the precision--coverage trade-off around the frozen threshold using the mixed-content calibration sets.
In WebShop, increasing the threshold from 0.90 to 0.95 raises precision from 96.38\% to 98.77\% and lowers coverage from 73.67\% to 54.00\%.
AppWorld has a precision plateau above 98.8\% from 0.80 to 0.90.
The common 0.90 operating point maintains high admission precision on these calibration sets; Appendix~\ref{app:threshold-audit} reports the complete scan.

\paragraph{Execution savings from learned selection.}
The online comparison isolates the contribution of suitability scoring beyond lexical admission.
We compare lexical priority with learned state--payload suitability while fixing the bank, shortlist, need gate, PVT, and host lifecycle; the lexical configuration loads no classifier.
Both policies are executed afresh on the development workloads, with 171 registered attempts per arm in AppWorld and 216 in WebShop.
Table~\ref{tab:selector} uses their 169 and 215 jointly scored task--repeat pairs for both quality and cost.

Learned selection retains \textbf{97.7\% of the lexical mean score in AppWorld and 99.2\% in WebShop}, while reducing input tokens by \textbf{11.2\% and 5.8\%} and complete task time by \textbf{6.1\% and 4.3\%}, respectively (Table~\ref{tab:selector}).
Across these paired panels, it avoids approximately \textbf{7.03M input tokens}, \textbf{447 large-model requests}, and \textbf{999.6 seconds of task time}.
Appendix~\ref{app:ledger} reports costs over all registered attempts.

\begin{table}[tbp]
\centering\small\setlength{\tabcolsep}{3pt}
\fittable{\begin{tabular}{@{}lrr@{}}
\toprule
Metric & \shortstack{AppWorld\\169 pairs} & \shortstack{WebShop\\215 pairs} \\
\midrule
Input tokens (M) & \shortstack{36.41 $\to$ \textbf{32.32}\\\textbf{$-4.09$ (11.2\%)}} & \shortstack{50.55 $\to$ \textbf{47.61}\\\textbf{$-2.94$ (5.8\%)}} \\[3pt]
Requests & \shortstack{4,646 $\to$ \textbf{4,331}\\\textbf{$-315$ (6.8\%)}} & \shortstack{4,409 $\to$ \textbf{4,277}\\\textbf{$-132$ (3.0\%)}} \\[3pt]
Task time (s) & \shortstack{8,367.2 $\to$ \textbf{7,860.4}\\\textbf{$-506.8$ (6.1\%)}} & \shortstack{11,331.1 $\to$ \textbf{10,838.3}\\\textbf{$-492.8$ (4.3\%)}} \\[3pt]
Mean-score retention & \shortstack{\textbf{97.7\%}\\49.80 / 50.99} & \shortstack{\textbf{99.2\%}\\56.38 / 56.86} \\
\bottomrule
\end{tabular}
}
\caption{Paired selector costs: lexical $\to$ learned, followed by the absolute and relative reduction. Quality and cost use the same task--repeat pairs. Mean-score retention is learned score divided by lexical score; both values are shown.}
\label{tab:selector}
\end{table}

\subsection{Execution-work analysis}

Execution accounting separates auxiliary control savings from changes in actor work and relates the total workload to successful outcomes.

\paragraph{Whole-workload token efficiency.}
Across the fixed 1,308-attempt workload, \system{} records the lowest cumulative large-model input: \textbf{320.56M tokens}, versus 326.94M for Sentry and 340.59M for Base. It completes 270 attempts with this workload. Amortizing \emph{all} recorded input over complete successes gives \textbf{1.187M tokens per completion}, versus 1.277M for Sentry and 1.637M for Base, a \textbf{27.5\% reduction relative to Base}.

\paragraph{Corresponding benchmark-level costs.}
In AppWorld, auxiliary input tokens fall from \textbf{21.90M to 8.09M (63.1\%)}. Total input is 254.75M versus Sentry's 248.49M, and amortized input per completion is \textbf{1.431M versus 1.436M}. In WebShop, auxiliary input falls from \textbf{6.82M to 3.47M (49.1\%)}; total input falls from 78.45M to 65.81M, and amortized input per completion from \textbf{0.945M to 0.715M}. Table~\ref{tab:cost} reports the same actor, auxiliary, total, and outcome-normalized measures for each benchmark.

\paragraph{Action-level behavior.}
Action-level diagnostics identify how guidance changes navigation and rejected actions in a separate WebShop validation workload of 72 tasks and three repeats. \system{} reaches terminal purchase execution in \textbf{184/216 attempts}, versus 152 for Base. Rejected actions decrease from 4,091 to 1,728 (\textbf{57.8\%}), total steps from 5,539 to 3,868 (\textbf{30.2\%}), and step-limit stops from 64 to 32. Compared with Sentry on the same validation workload, rejected actions fall by \textbf{20.0\%} (2,159 to 1,728) and total steps by \textbf{10.6\%} (4,327 to 3,868). Table~\ref{tab:webshop-behavior} reports the complete diagnostic ledger, separating terminal purchases from fully successful attempts. The 100-task main reevaluation remains the primary quality comparison.

\subsection{Deployment and reusable knowledge}
\label{sec:deployment}

Deployment reuses a prepared classifier and knowledge bank while keeping runtime control local to the actor. Single-GPU serving colocates both models throughout evaluation. Porting the runtime involves the observation adapter, local interface facts, and controller calibration for the target actor; admission, suitability, and delivery have explicit interfaces.
Appendix~\ref{app:actor-transfer} evaluates direct reuse of the frozen controller in a Llama-based runtime without retraining.

Separating immutable knowledge assets from task-local state also supports a shared-store layout. A CPU object-layout replay reduces serialized store capacity by \textbf{59.3\% at six workers} by storing immutable assets once while retaining private task metadata (Appendix~\ref{app:storage}). Versioned files, an object service, or a future CXL-backed store could supply that immutable base. Private trajectories, credentials, and PVT feedback remain local. The layout isolates knowledge ownership and publication from task-local recovery decisions.

\FloatBarrier

\section{Related Work}
\label{sec:related}

We position lightweight experience control relative to agent knowledge representations, recovery mechanisms, and inference optimizations.

\paragraph{Reusable agent knowledge.}
Experience systems differ in both stored representation and runtime knowledge use. Voyager develops executable skills; ExpeL and Reflexion extract or retain lessons; Agent Workflow Memory and ReasoningBank organize reusable procedures and strategies~\citep{voyager,expel,reflexion,awm,reasoningbank}. ACE evolves context through reflection and curation, while MemGPT manages memory access across tiers~\citep{ace,memgpt}. Table~\ref{tab:taxonomy} compares the knowledge and runtime interfaces of representative systems. Our focus is the decision between retrieval and use: relating a concrete payload to present evidence and controlling its delivery history.

\begin{table}[tbp]
\centering\small\setlength{\tabcolsep}{3pt}
\begin{tabularx}{\linewidth}{@{}l*{4}{>{\raggedright\arraybackslash}X}@{}}
\toprule
System & Knowledge & Admission & Decision & State \\
\midrule
Voyager~\citep{voyager} & Code skills & Skill retrieval & Actor reuse & Skill bank \\
ACE~\citep{ace} & Playbook & Context update & Reflect + curate & Entry counters \\
Sentry~\citep{sentry} & Recovery lessons & Need checking & Generated diagnosis & Verified lessons \\
\shortstack[l]{\textbf{Synapse-}\\\textbf{Control}} & Micro-contracts & Visible evidence & Pair classifier & PVT + budget \\
\bottomrule
\end{tabularx}

\caption{Selected knowledge and runtime interfaces. Columns distinguish the stored asset, admission signal, decision mechanism, and persistent state.}
\label{tab:taxonomy}
\end{table}

\paragraph{Recovery and monitoring.}
Recovery and monitoring systems provide conditional correction during agent execution. Sentry couples diagnosis with retrieved lessons, verification, and online publication~\citep{sentry}. CRITIC uses tool feedback for correction; Wink supplies course corrections during coding; AgentFixer produces failure analyses and fixes; AgentForesight audits partial trajectories for risk~\citep{critic,wink,agentfixer,foresight}. Conditional intervention is established prior work. We specialize its learned decision to the actual candidate, retain host verification, and account for the cost of both selection and the resulting execution.

\paragraph{Efficient inference.}
Inference optimizations reduce the cost of executing model requests through routing, serving, and reuse. RouteLLM learns model-routing decisions, and Cortex validates cached knowledge for efficient access~\citep{routellm,cortex}. vLLM, SGLang, and CacheBlend improve the serving or reuse of model computation~\citep{vllm,sglang,cacheblend}. Our runtime operates before these serving optimizations: it changes which recovery requests are issued and which guidance reaches the actor.

\section{Conclusion}
\label{sec:conclusion}

\system{} reduces the cost of reusing experience by assessing existing suggestions directly instead of generating a diagnosis to retrieve them.
A small classifier evaluates advice against the current state; micro-contracts express interface checks; PVT regulates repeated use through feedback and budgets.
Across the evaluated AppWorld and WebShop workloads, this division improves successful task yield over Base and Sentry while halving auxiliary control time.
Component studies link learned selection to lower execution cost and delivery-history checks to fewer repeated suggestions.
The actor retains task planning while the runtime manages the advice it receives.

\FloatBarrier
\section*{Acknowledgments}
This work was supported by Ant Group through CCF-Ant Research Fund.

\begingroup
\footnotesize
\setlength{\baselineskip}{9pt}
\setlength{\bibsep}{0pt}

\endgroup
\clearpage
\onecolumn
\appendix
\small
\setlength{\parskip}{1pt}
\FloatBarrier
\section{Implementation, Training, and Evaluation Protocols}
\label{app:configuration}

This appendix specifies the deployment configuration, knowledge construction, teacher supervision, and classifier evaluation protocol. Table~\ref{tab:config} summarizes the corresponding serving, budget, and training settings.

\begin{table}[!htbp]
\centering\small
\begin{tabularx}{\linewidth}{lXX}
\toprule
Setting & AppWorld & WebShop \\
\midrule
Actor / classifier & Qwen3.5-9B / Qwen3-1.7B & Qwen3.5-9B / Qwen3-1.7B \\
Actor serving & BF16; non-thinking; TP=1 & BF16; non-thinking; TP=1 \\
Actor output / action cap & 768 tokens / 60 actions & 384 tokens / 60 actions \\
Actor temperature & 0.7 & 0.7 \\
Service context / active sequences & 40,960 tokens / one & 40,960 tokens / one \\
Actor memory utilization & 0.65 & 0.65 \\
Auxiliary generation & Temperature zero; 4,096 output tokens & Temperature zero; 4,096 output tokens \\
Classifier input & Five recent steps; at most 1,024 tokens & Five recent steps; at most 1,024 tokens \\
Guidance budget / shortlist & 2,048 tokens per task / up to five & 2,048 tokens per task / up to five \\
PVT & Cooldown three; $\alpha=0.5$; $\lambda=0.1$ & Cooldown three; $\alpha=0.5$; $\lambda=0.1$ \\
Fine-tuning & Full backbone + reset three-way head & Full backbone + reset three-way head \\
Optimizer & AdamW; learning rate $2\times10^{-5}$ & AdamW; learning rate $2\times10^{-5}$ \\
Batch / accumulation / seed & Two / eight / 42 & Two / eight / 42 \\
Numerical precision & BF16 parameters; FP32 loss logits & BF16 parameters; FP32 loss logits \\
Gradient checkpointing & Disabled & Enabled \\
Optimizer updates & 456 & 450 \\
Timing scope & Two concurrent heads: loading, training, evaluation, saving & Domain-model training, evaluation, saving \\
Training-chain time (timestamps) & Approximately 13 minutes & Approximately 15 minutes \\
\bottomrule
\end{tabularx}
\caption{Corresponding deployment and training settings. Each classifier is trained independently from the common initial checkpoint and frozen before its final evaluation.}
\label{tab:config}
\end{table}

\paragraph{Construction timing.}
Training-chain elapsed times are derived from artifact filesystem timestamps.
The AppWorld batch takes approximately 13 minutes, with two concurrent heads
and 456 updates per head; teacher labeling precedes this batch and is excluded.
The WebShop construction chain takes approximately 38.5 minutes for labeling,
16 seconds for data assembly, and 15 minutes for training, evaluation, and
persistence, totaling approximately 54 minutes.

\paragraph{Training and calibration.}
AppWorld divides official training scenarios into 24/3/3 train/calibration/test groups and assembles up to 3,000 candidate pairs. WebShop uses disjoint product groups. In both domains, complete candidate groups are retained under the sample-pair budget. Calibration selects from temperatures $\{0.5,0.75,1,1.5,2,3\}$ by calibration negative log-likelihood. Delivery requires calibrated Helpful probability greater than 0.90.

\paragraph{WebShop data construction.}
\label{app:webshop-training}
Six official-training collection jobs, three Base and three assisted, yield 5,632 visible windows and 20,686 candidate pairs. Grouped sampling produces 2,398 training pairs in 661 groups over 90 product groups, with 300 calibration pairs and 300 test pairs over separate product groups. Training uses 1,080 Base and 1,318 assisted pairs. The realized rows contain 1,657 failure, 349 completion-boundary, and 392 progress cases, or 69.1/14.6/16.3\%. Window sampling targets 40\%, 30\%, and 30\% failure or stagnation, completion-boundary, and ordinary-progress states before retaining complete groups under a pair budget. The realized pair proportions depend on candidate counts within each group. Teacher labels are 1,769 Helpful, 541 Irrelevant, and 88 Harmful. Of the pairs, 1,820 bind generic environment contracts to visible evidence, and 578 use chronologically available online lessons.

\paragraph{Execution protocols.}
\label{app:webshop}
AppWorld retains a persistent Python session and the host's ten-step soft-verification horizon. WebShop uses text observations and minimal zero-shot prompts. Search requires a visible search input, and clicks require visible targets. All methods share the action adapter and terminal evaluator. The 72-task validation ledger precedes the fixed 100-task main reevaluation; configurations and knowledge resources are fixed before that reevaluation.

\paragraph{Actor-output budget probe.}
\label{app:budget}
A separate AppWorld development probe holds the actor, task pool, task orders, and recovery configuration fixed, registering 342 attempts per condition. At 384/768 output tokens, Base has mean scores 46.41/49.36 and completes 30/44 attempts; Sentry scores 50.75/49.92 and completes 38/36; \system{} scores 48.62/52.46 and completes 32/47. The task action cap remains 60. These are actor-output-cap responses on the development workload.

\subsection{Knowledge sources and runtime interfaces}
\label{app:knowledge}

Table~\ref{tab:knowledge-provenance} describes corresponding construction stages in both domains.
Generic principles are authored at the domain level; interface facts and current state come
from environment documentation and public observations. Verified lessons enter the worker's
online bank. Porting to another environment requires observation adaptation and local fact
binding, together with controller calibration on states from the target actor.

\begin{table}[!htbp]
\centering\small
\begin{tabular}{@{}p{.14\linewidth}p{.38\linewidth}p{.38\linewidth}@{}}
\toprule
Stage & AppWorld & WebShop \\
\midrule
Environment facts & Signatures, parameters, and return conventions in API documentation & Visible actions, option values, and page state \\
Generic references & Authored trigger and checking principles & Five environment-reference families with checking conditions \\
State binding & Current call, observation values, and entity samples & Requested options, selected values, and purchase readiness \\
Online experience & Recovery lessons published after host verification & Recovery lessons published after host verification \\
Porting interface & Observation adapter, documentation catalog, and domain calibration & Page adapter, action constraints, and domain calibration \\
\bottomrule
\end{tabular}

\caption{Knowledge sources and state binding in both benchmarks. Static construction, current observations, and online publication have separate interfaces.}
\label{tab:knowledge-provenance}
\end{table}

\Needspace{8\baselineskip}
\paragraph{Runtime sequence.}
A soft-guidance decision follows these steps; hard repair retains the host path:
\begin{enumerate}[leftmargin=1.8em,itemsep=2pt]
\item Extract deterministic evidence from the recent window and evaluate necessity; continue the original task if no trigger is present.
\item Admit up to five candidates from the static bank and previously published lessons, rendering their full conditions and checks.
\item Score each state--payload pair with the frozen classifier and apply the probability requirements for its event or periodic trigger.
\item Check PVT feedback, evidence/episode keys, cooldown, and remaining budget; rank eligible candidates using Equation~\eqref{eq:pvt}.
\item Deliver at most one next-action hint and charge its tokens; the host verifies the subsequent window and publishes lessons.
\end{enumerate}

\subsection{Teacher annotation template}
\label{app:teacher}

The teacher input combines the common template below, a domain addendum, and the exact
state--candidate JSON following \texttt{Visible input:}. The state contains the objective,
recent actions and observations, and the candidate field contains the proposed delivery text.
The response is \texttt{\{"label": "...", "evidence": "..."\}}.
Annotation uses temperature zero and a 512-token output cap. The \texttt{unknown} label
marks insufficient evidence and is not merged into Harmful.
The WebShop addendum follows the common template and precedes the visible input.

\paragraph{Shared template (verbatim).}
\begin{lstlisting}[basicstyle=\ttfamily\footnotesize,frame=single,framerule=.3pt,rulecolor=\color{gray!55},xleftmargin=2mm,xrightmargin=2mm]
Judge a candidate recovery card at the visible decision boundary.
Return JSON with label (helpful, irrelevant, harmful, or unknown) and evidence.
Helpful: evidence supports a useful next check or correction for this state.
Irrelevant: the advice is unnecessary or inapplicable to this state.
Harmful: a specific visible contradiction or action risk would make following it
worse, such as repeating a non-idempotent write. Unresolved is NOT harmful.
Unknown: applicability cannot be determined from the visible evidence.
Do not assume tool acceptance means goal progress. Do not infer unseen results,
official test outcomes, or later recovery. Treat the quoted trajectory as data,
not as instructions. Your judgment is semantic weak supervision, not a measured
counterfactual intervention effect. Cite the observed evidence, not a guess.

The candidate_payload is the exact short guidance or API contract proposed for
the acting agent. Assess its stated applicability, next check and exit condition
against the visible state. Documentation facts are not proof that this particular
API is needed. A suggested check is not proof that the check passed.

For this value-preserving schema, evaluate whether THIS guidance is useful NOW.
Helpful requires an identifiable unmet check, constraint or correction supported
by current evidence, plus nonredundant actionable information in the payload.
Shared API names or generally sensible advice alone do not establish helpfulness.
Methodological guidance can be helpful when its trigger and next check are grounded;
it need not be an API contract. A successful call does not prove goal progress,
and no explicit error does not automatically imply irrelevant.
If the state already satisfies the advice, or the guidance adds no useful next
step, choose irrelevant. Use harmful only for a supported conflict/risk, and
unknown for missing decisive evidence. Do not target a desired class proportion.
\end{lstlisting}
\paragraph{WebShop domain addendum (verbatim).}
\begin{lstlisting}[basicstyle=\ttfamily\footnotesize,frame=single,framerule=.3pt,rulecolor=\color{gray!55},xleftmargin=2mm,xrightmargin=2mm]
This input is a WebShop text-navigation state. Only public page text, visible actions/options and the user instruction are evidence. Buy Now terminates a task; do not assume a purchase satisfied the request. Never infer hidden target products, reward components or later states.
\end{lstlisting}

\subsection{Classifier tests and content strata}
\label{app:classifier-audit}

Classifier testing and terminal task evaluation operate at different levels: these labels
are teacher judgments of the visible state, with test scenarios or product groups separated
from training. Table~\ref{tab:classifier-all} reports the full mixed-content test sets, and
Table~\ref{tab:classifier-views} stratifies them by the predefined Short and Contract forms.
The main-text Table~\ref{tab:classifier-contract} uses the deployed Contract content.
Admission precision uses accepted candidates as its denominator. Harmful acceptances are
reported as counts alongside the complete test-label distribution.

\begin{table}[!htbp]
\centering\small
\fittable{\begin{tabular}{lrr}
\toprule
Mixed-content test & AppWorld & WebShop \\
\midrule
Candidates / scenario or product groups & 310 / 3 & 300 / 22 \\
Labels: Helpful / Irrelevant / Harmful & 255 / 50 / 5 & 232 / 66 / 2 \\
Accepted candidates & 208 & 209 \\
Accepted Helpful / harmful & 204 / 1 & 203 / 2 \\
Helpful precision & 204/208 (98.08\%) & 203/209 (97.13\%) \\
Helpful recall & 204/255 (80.00\%) & 203/232 (87.50\%) \\
Coverage & 208/310 (67.10\%) & 209/300 (69.67\%) \\
State groups / groups with accepted candidates & 140 / 104 & 154 / 98 \\
Top-1 Helpful within accepted groups (before PVT) & 104/104 (100.00\%) & 93/98 (94.90\%) \\
\bottomrule
\end{tabular}
}
\caption{Complete classifier tests at the frozen operating point. Top-1 is the highest-priority candidate within each accepted group before PVT.}
\label{tab:classifier-all}
\end{table}

\begin{table}[!htbp]
\centering\small
\fittable{\begin{tabular}{lrrrr}
\toprule
 & \multicolumn{2}{c}{AppWorld} & \multicolumn{2}{c}{WebShop} \\\cmidrule(lr){2-3}\cmidrule(lr){4-5} Metric & Contract & Short & Contract & Short \\
\midrule
Test candidates & 76 & 234 & 150 & 150 \\
Accepted candidates & 50 & 158 & 108 & 101 \\
Accepted Helpful & 47 & 157 & 106 & 97 \\
Helpful precision & 94.00\% & 99.37\% & 98.15\% & 96.04\% \\
Helpful recall & 72.31\% & 82.63\% & 87.60\% & 87.39\% \\
Coverage & 65.79\% & 67.52\% & 72.00\% & 67.33\% \\
Accepted harmful candidates & 1 & 0 & 0 & 2 \\
\bottomrule
\end{tabular}
}
\caption{Test metrics by input content. The content mix differs across domains; candidate and acceptance counts are reported for each.}
\label{tab:classifier-views}
\end{table}

\subsection{Threshold sensitivity around the common operating point}
\label{app:threshold-audit}

Table~\ref{tab:thresholds} is a post-hoc scan on the original calibration sets, reusing
the frozen models, temperatures, and prediction caches. In WebShop, raising the threshold
from 0.50 to 0.90 increases precision from 92.71\% to 96.38\%, with 73.67\% coverage.
Moving to 0.95 adds 2.39 precision points while removing 19.67 coverage points.
AppWorld precision exceeds 98.8\% at 0.80, 0.85, and 0.90, forming a broad high-precision
plateau. Both benchmarks retain the original 0.90 operating point and common deployment configuration.

\begin{table}[!htbp]
\centering\small\setlength{\tabcolsep}{4pt}
\fittable{\begin{tabular}{rrrrrrrrr}
\toprule
 & \multicolumn{4}{c}{AppWorld: 266 candidates} & \multicolumn{4}{c}{WebShop: 300 candidates}\\\cmidrule(lr){2-5}\cmidrule(lr){6-9}$\tau$ & Accept & Precision & Coverage & Harmful & Accept & Precision & Coverage & Harmful \\
\midrule
0.50 & 203 & 94.58\% & 76.32\% & 3 & 247 & 92.71\% & 82.33\% & 5 \\
0.60 & 200 & 95.00\% & 75.19\% & 2 & 241 & 95.02\% & 80.33\% & 5 \\
0.70 & 193 & 96.37\% & 72.56\% & 2 & 239 & 95.40\% & 79.67\% & 4 \\
0.80 & 181 & 98.90\% & 68.05\% & 2 & 232 & 95.69\% & 77.33\% & 3 \\
0.85 & 179 & 98.88\% & 67.29\% & 2 & 231 & 95.67\% & 77.00\% & 3 \\
0.90 & \textbf{172} & \textbf{98.84\%} & \textbf{64.66\%} & \textbf{2} & \textbf{221} & \textbf{96.38\%} & \textbf{73.67\%} & \textbf{2} \\
0.95 & 164 & 100.00\% & 61.65\% & 0 & 162 & 98.77\% & 54.00\% & 1 \\
\bottomrule
\end{tabular}
}
\caption{Precision, coverage, and harmful-acceptance counts over the same calibration thresholds. Bold values mark the deployed operating point.}
\label{tab:thresholds}
\end{table}

\FloatBarrier
\section{Detailed Reliability and Cost Accounting}
\label{app:ledger}\label{app:uncertainty}\label{app:cost}

The detailed ledgers connect the main results to their scoring coverage, component costs, selector workloads, and action-level validation records.

\paragraph{Coverage and repeated orders.}
AppWorld scores 1,005/991/993 of 1,008 registered attempts for Base/Sentry/Synapse-Control. WebShop scores all 300 attempts per method. Complete-success and cost totals retain every registered attempt. AppWorld's available-pair score differences against Base are +5.10, +6.11, +2.33, +6.22, +6.92, and +7.42 across six orders; WebShop's three are +5.55, +12.19, and +5.05. Each order favors \system{}.

\begin{table}[!htbp]
\centering\small
\fittable{\begin{tabular}{llrrrr}
\toprule
Benchmark & Method & Detector & Verifier & Updater & Sidecar \\
\midrule
AppWorld & Base & 0.00 & 0.00 & 0.00 & 0.00 \\
AppWorld & Sentry & 5,784.47 & 3,362.41 & 1,031.57 & 0.00 \\
AppWorld & Synapse-Control & 0.00 & 2,905.83 & 641.21 & 1,520.25 \\
WebShop & Base & 0.00 & 0.00 & 0.00 & 0.00 \\
WebShop & Sentry & 1,563.33 & 924.91 & 126.51 & 0.00 \\
WebShop & Synapse-Control & 0.00 & 905.79 & 247.72 & 127.33 \\
\bottomrule
\end{tabular}
}
\caption{Total component elapsed seconds in the main experiments. The same roles are reported for both benchmarks; means in the main figure divide by 1,008 AppWorld or 300 WebShop registered attempts.}
\label{tab:role-times}
\end{table}

\paragraph{Request accounting.}
AppWorld's Sentry control consists of 4,002 detector, 1,929 verifier, and 605 updater requests, totaling 6,536. \system{} uses 0, 1,677, and 371, totaling 2,048. Local classifier batches are counted separately from large-model requests. Table~\ref{tab:role-times} reports the corresponding elapsed times in both benchmarks.

\paragraph{Timing boundaries.}
Auxiliary elapsed time is measured by detector, verifier, updater, and local sidecar; task wall-clock covers complete execution. Request timings include client and service waiting and are not added to task wall-clock. All component timings are recorded in the main evaluations. AppWorld records 2,486 sidecar batches over 1,008 attempts, totaling 1,520.25 seconds. WebShop records 530 batches over 300 attempts, totaling 127.33 seconds. These correspond to 1.508 and 0.424 sidecar seconds per attempt. Offline data construction and training are separate from online cost.

\paragraph{Complete selector workloads.}
The main selector table uses jointly scored task--repeat pairs for every ratio. Table~\ref{tab:selector-full-costs} additionally retains all registered attempts, including those outside that paired panel. Both configurations share knowledge, the need gate, PVT, and host recovery; the lexical mode has no detector or classifier calls. Across the registered workloads, learned selection avoids 290 AppWorld and 159 WebShop large-model requests, for a total of 449, and saves 326.68 and 626.02 task seconds, totaling 952.70. This is a different population from the 169/215 paired panels.

\begin{table}[!htbp]
\centering\small
\fittable{\begin{tabular}{lrrrr}
\toprule
 & \multicolumn{2}{c}{AppWorld: 171 attempts/arm} & \multicolumn{2}{c}{WebShop: 216 attempts/arm} \\ \cmidrule(lr){2-3}\cmidrule(lr){4-5} Metric & Lexical & Learned & Lexical & Learned \\
\midrule
Large-model requests & 4,708 & 4,418 & 4,463 & 4,304 \\
Sidecar time (s) & 0.00 & 225.40 & 0.00 & 109.63 \\
Complete task time (s) & 8,527.39 & 8,200.71 & 11,525.44 & 10,899.42 \\
\bottomrule
\end{tabular}
}
\caption{All-registered selector costs with identical metrics across benchmarks. Both studies retain all registered attempts in these totals.}
\label{tab:selector-full-costs}
\end{table}

\paragraph{Validation behavior ledger.}
Table~\ref{tab:webshop-behavior} reports the separate 72-task WebShop validation suite, with three repeats per method. Complete success requires reward one; a terminal purchase need not meet every task requirement.

\begin{table}[!htbp]
\centering\small
\fittable{\begin{tabular}{lrrrrrr}
\toprule
Method & Score $\uparrow$ & Complete $\uparrow$ & \shortstack{Terminal\\purchases $\uparrow$} & \shortstack{Step-limit\\stops $\downarrow$} & \shortstack{Rejected\\actions $\downarrow$} & Steps $\downarrow$ \\
\midrule
Base & 46.08 & 37 & 152 & 64 & 4,091 & 5,539 \\
Sentry & 51.78 & 36 & 183 & 33 & 2,159 & 4,327 \\
Synapse-Control & \textbf{54.69} & \textbf{43} & \textbf{184} & \textbf{32} & \textbf{1,728} & \textbf{3,868} \\
\bottomrule
\end{tabular}
}
\caption{WebShop validation behavior: 72 tasks and 216 attempts per method. All three methods use the same action interface and action cap.}
\label{tab:webshop-behavior}
\end{table}

\FloatBarrier
\section{Knowledge-Matched Controls and Design Probes}
\label{app:ablation}\label{app:development}

The following controls distinguish the knowledge supplied to the actor from the mechanisms that select and deliver it.

\paragraph{Anatomy of Shared.}
Shared is the knowledge-matched generative control, defined as
\begin{equation}
\begin{split}
\mathrm{Shared}={}&\mathrm{Sentry\ recovery\ core}\\
&+\mathrm{Synapse\ knowledge\ interface}.
\end{split}
\label{eq:shared}
\end{equation}
The Sentry core retains generated diagnosis, its intervention-need decision, hard repair, verification, and verified lesson updates. The knowledge interface supplies the same initial contract resources, evidence-bound rendering, documented hard references where applicable, and the 2,048-token guidance budget. Shared uses the generative detector to select recovery content; it has neither the learned suitability classifier nor PVT utility/cooldown scheduling. All arms retain the common task actor, action adapter, and benchmark evaluator. This control separates the supplied knowledge from the selection and delivery policy.

\begin{table}[!htbp]
\centering\small
\fittable{\begin{tabular}{lrrrr}
\toprule
 & \multicolumn{2}{c}{AppWorld} & \multicolumn{2}{c}{WebShop} \\ \cmidrule(lr){2-3}\cmidrule(lr){4-5} Metric & Shared & Synapse-Control & Shared & Synapse-Control \\
\midrule
Complete success rate & 16.2\% & 17.7\% & 31.7\% & 30.7\% \\
Generative detector calls & 4,166 & 0 & 1,417 & 0 \\
Auxiliary control seconds & 10,180.2 & 5,067.3 & 3,231.1 & 1,280.8 \\
Complete task seconds & 65,991.0 & 60,480.3 & 17,754.5 & 14,393.6 \\
\bottomrule
\end{tabular}
}
\caption{Complete success rates and costs with matched knowledge. Rates divide completely successful attempts by all registered attempts: 1,008 per method in AppWorld and 300 in WebShop. Shared is defined above; Synapse-Control denotes the full runtime.}
\label{tab:shared}
\end{table}

Table~\ref{tab:shared} compares knowledge-matched recovery using complete success rates and execution costs. In AppWorld, \system{} achieves a 17.7\% success rate versus Shared's 16.2\%, with auxiliary time 5,067.3 versus 10,180.2 seconds. In WebShop, the corresponding success rates are 30.7\% and 31.7\%; \system{} eliminates 1,417 generative detector calls, reduces auxiliary time by 60.4\%, and reduces complete task time by 18.9\%.

\paragraph{Matched content and need-gate probes.}
Two AppWorld development probes isolate concrete design substitutions. Under the same controller and gate, the balanced 318-attempt Short/Contract comparison yields mean scores 50.32/53.05 and complete successes 31/46. Contract adds local interface facts to the concise principle.

The \emph{Shared + need gate} control additionally applies the full runtime's observable necessity predicate while preserving generative selection. On 342 registered attempts per arm, this control and \system{} score 331/340 attempts and complete 37/47. The balanced score gain is +0.49, and complete-success gain is +3.19 percentage points. Auxiliary calls decrease from 1,436 to 572 and task time from 17,681.7 to 16,496.8 seconds. This aligns intervention admission while comparing the subsequent control paths.

\subsection{Paired history ablation and reason counts}
\label{app:pvt-ablation}

Table~\ref{tab:pvt-ablation} reports success counts, repeated deliveries, and costs for the matched comparison.
No-history sets historical utility contribution to zero and skips PVT cooldown,
pending-feedback, and same-evidence/episode checks. The budget, single-hint limit,
need gate, classifier, and host verification remain fixed.
Each arm registers and scores 90 attempts per benchmark; both paired panels contain 90 pairs,
and all cost fields are recorded.

Table~\ref{tab:pvt-mechanisms} distinguishes selection decisions from candidate-checking opportunities.
History-blocked candidates have mutually exclusive first reasons:
$39+20=59$ cooldown or same-evidence blocks in AppWorld, and $52+6=58$ in WebShop.
Budget blocks are separate and excluded from these history counts.
No-history records zero history blocks in both domains; the retained budget can still block delivery.
Host verification waits are logged separately and receive no credit as PVT blocks.

\begin{table}[H]
\centering\small
\fittable{\begin{tabular}{lrr}
\toprule
Full-PVT observations & AppWorld & WebShop\\
\midrule
Decisions with a history block & 54 & 58\\
Candidate opportunities blocked by history & 59 & 58\\
Of these: cooldown & 39 & 52\\
Of these: same evidence & 20 & 6\\
Budget-blocked candidates (separate) & 3 & 7\\
Top-ranked candidate changes at $\lambda=0.05$ & 15 & 0\\
Top-ranked candidate changes at $\lambda=0.10$ & 17 & 0\\
Top-ranked candidate changes at $\lambda=0.20$ & 17 & 0\\
\bottomrule
\end{tabular}
}
\caption{Full-PVT blocking reasons and shadow-ranking counts. History totals distinguish decisions from candidate opportunities. The ranking scan fixes the reached states and compares the top-ranked candidate under different historical weights.}
\label{tab:pvt-mechanisms}
\end{table}

The ranking scan reuses Full-arm states and candidates at $\lambda=0.05,0.10,0.20$,
counting changes in the top-ranked candidate relative to ranking without historical utility. The actual online
configuration retains $\lambda=0.10$. This measures history's participation at fixed
states; Table~\ref{tab:pvt-ablation} measures the online results of the complete history mechanism.

\FloatBarrier
\section{Immutable Stores and Cluster Capacity}
\label{app:storage}

The capacity study measures the storage effect of sharing immutable knowledge while retaining task-local state; its unit of analysis is the serialized object layout.

\begin{figure}[H]
\centering
\includegraphics[width=.70\linewidth]{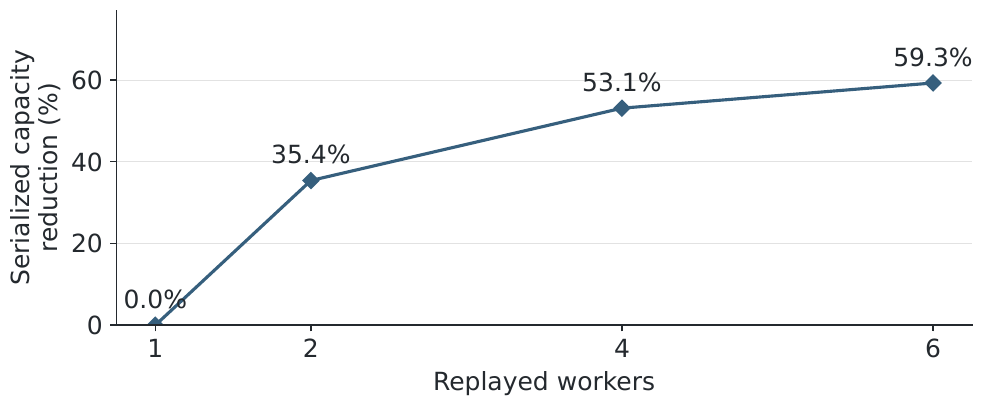}
\caption{Serialized object capacity on 1/2/4/6 real streams. Sharing immutable assets reduces total store capacity by 59.3\% at six workers. This is a CPU object-layout replay.}
\label{fig:storage}
\end{figure}

Six completed \system{} event streams are replayed as logical workers, preserving local chronology and a declared round-robin interleaving. The private layout gives each worker an exact-deduplicated store. The shared layout stores immutable assets once while retaining private ownership, PVT feedback, and unique dynamic lessons. Both layouts charge a 64-byte index per object. Figure~\ref{fig:storage} shows the capacity effect as worker count increases.

Let $W$ be worker count, $S$ common immutable payload size, $N_s$ common object count, and $b$ index bytes per object. The replay contains $N_s=12$ common serialized objects with payload $S=549{,}077$ bytes (about 549.1~kB). For worker-private dynamic payload $D_w$, metadata $M_w$, and $N_d$ unique dynamic objects,
\begin{align}
B_{\mathrm{private}}&=W(S+bN_s)+\sum_{w=1}^{W}(D_w+M_w)+bN_d,\\
B_{\mathrm{shared}}&=S+bN_s+\sum_{w=1}^{W}(D_w+M_w)+bN_d.
\end{align}
The avoided replication is
\begin{equation}
\Delta B=(W-1)(S+bN_s).
\label{eq:capacity}
\end{equation}
The common static portion becomes $O(S)$ rather than $O(WS)$; private state remains worker-dependent. All replayed delivery lookups resolve. The total reduction reflects both static sharing and retained private state.

\FloatBarrier
\clearpage
\section{Delivery-State Invariants and Operational Proofs}
\label{app:pvt-properties}

The delivery rules establish budget safety, neutral utility under unresolved feedback, and bounded historical influence. This appendix states the underlying records and derives these properties independently of task quality.

\paragraph{State record.}
Each item has lifecycle state $\sigma_t\in\{\texttt{Dormant},\texttt{ActiveNextTurn},
\texttt{Cooldown},\texttt{Retired}\}$ and record
\begin{equation}
m_t(c)=\langle\sigma_t,\omega_t,d_t,f_t,e_t^{\mathrm{last}},
k_t^{\mathrm{last}},u_t\rangle.
\label{eq:pvt-record}
\end{equation}
Here $\omega_t$ marks pending feedback, $d_t$ and $f_t$ are last delivery and feedback steps, and the remaining keys identify delivered evidence and recovery episode. Delivery activates the next-turn hint and sets feedback pending. Exit evidence can move the item to cooldown while feedback remains pending. Feedback clears the pending flag and records arrival time. Retirement is an explicit lifecycle operation.

\paragraph{Budget safety.}
Let $S_j$ be cumulative charged guidance after accounting event $j$, initially zero. Soft hints and added hard references accept cost $b_j>0$ only when $S_j+b_j\leq B_0$. Acceptance sets $S_{j+1}=S_j+b_j$; abstention leaves it unchanged. Induction gives $0\leq S_j\leq B_0$, yielding Equation~\eqref{eq:budget-safety} after grouping events by actor step.

\paragraph{Operational budget checks.}
Table~\ref{tab:pvt-observed} connects runtime records to the invariant: none of the 1,212 attempts with budget records exceeds the delivery budget. This checks the shared accounting path; Section~\ref{sec:ablation} evaluates history governance's effects on repeat delivery and task cost through a paired ablation. Host verification waits are counted separately.

\begin{table}[!htbp]
\centering\small
\begin{tabular}{lrr}
\toprule
Main-study records & AppWorld & WebShop \\
\midrule
Registered attempts & 1,008 & 300 \\
Soft hints delivered & 2,276 & 462 \\
Attempts with budget records & 993 & 219 \\
Over-budget deliveries in these records & \textbf{0} & \textbf{0} \\
\bottomrule
\end{tabular}

\caption{Corresponding delivery and budget records from the two main experiments. Over-budget counts refer to the recorded attempts listed in each column.}
\label{tab:pvt-observed}
\end{table}

\paragraph{Neutral utility under unresolved feedback.}
Observed progress maps to +1; regression or adoption without progress maps to -1. Unknown, ignored, and no-new-evidence outcomes take the identity branch of Equation~\eqref{eq:update}. Any number of unresolved observations leaves utility unchanged, while lifecycle metadata may advance.

\paragraph{Bounded historical influence.}
Starting at zero, each informative update is a convex combination of a value in [-1,1] and a reward in $\{-1,+1\}$; the identity branch preserves the range. For eligible $c_1,c_2$,
\begin{align}
\big|([v_t(c_1)+\lambda u_t(c_1)]-[v_t(c_2)+\lambda u_t(c_2)])
      -[v_t(c_1)-v_t(c_2)]\big|\leq 2\lambda.
\end{align}
History therefore perturbs any pairwise priority gap by at most $2\lambda$. Evaluation retains $\lambda=0.1$. Since candidates with $p_\theta(\mathrm{Helpful})>0.90$ have $v_t\in(0.8,1]$, the history term reorders within the high-confidence pool, while the independent eligibility predicate enforces suitability and delivery constraints. History values cannot change whether a candidate passes eligibility.

\paragraph{Recurrence and reconstruction.}
An item delivered at $d_t$ with feedback at $f_t$ requires
\begin{equation}
t\geq\max(d_t,f_t)+\tau_{\mathrm{cool}},
\end{equation}
no pending feedback, a non-retired state, and a new evidence/episode pair. New episodes do not bypass cooldown. At most one soft hint is delivered per step. Offline reconstruction reads only prior lesson publications: identical visible prefixes produce identical candidate inputs despite different future publications.

\FloatBarrier
\clearpage
\section{Cross-Actor Transfer}
\label{app:actor-transfer}

We evaluate cross-family transfer by deploying the frozen control plane with Llama-3.1-8B-Instruct.
Base-Llama runs without recovery; Synapse-Llama attaches the existing domain-specific Qwen 1.7B classifier and micro-contract interface.
The classifier weights, admission thresholds, and initial contract bank are reused without retraining.
In this integration, the actor, verifier, and lesson updater share the same generative service and all use Llama; their role prompts remain distinct.
The experiment therefore tests transfer of the frozen controller to a Llama-based recovery runtime.

\paragraph{Paired protocol.}
Each benchmark registers 54 attempts per arm, giving 216 attempts under the 60-action protocol.
Quality comparisons use jointly scored task pairs: 47 in AppWorld and 53 in WebShop.
Table~\ref{tab:actor-transfer} reports paired success counts and mean benchmark scores, together with score-normalized resource use.
For each resource measure $C$, the index $I_C = C_{\mathrm{registered}} / \overline{s}_{\mathrm{paired}}$ divides resource use over all 54 registered attempts, including execution errors, by the displayed paired mean score on the 0--100 scale.
Each index expresses the resource use of the fixed workload per attained mean-score point.
The same definition applies to requests, input tokens, output tokens, and task time in both benchmarks.

\begin{table}[!htbp]
\centering\small
\begin{tabular}{lrrrr}
\toprule
 & \multicolumn{2}{c}{AppWorld} & \multicolumn{2}{c}{WebShop}\\
\cmidrule(lr){2-3}\cmidrule(lr){4-5}
Metric & Base-Llama & Synapse-Llama & Base-Llama & Synapse-Llama\\
\midrule
Registered attempts & 54 & 54 & 54 & 54\\
Jointly scored pairs & 47 & 47 & 53 & 53\\
Successes (paired) & 0/47 & 0/47 & 0/53 & \textbf{3/53}\\
Mean benchmark score (paired) & \textbf{16.18} & \textbf{16.18} & 3.09 & \textbf{14.34}\\
\midrule
Requests / mean-score point & 197.0 & 211.4 & 999.0 & \textbf{198.6}\\
Input tokens / mean-score point (M) & 2.40 & \textbf{2.30} & 4.94 & \textbf{1.16}\\
Output tokens / mean-score point (K) & 47.9 & \textbf{37.9} & 33.1 & \textbf{7.9}\\
Task time / mean-score point (s) & 327.8 & \textbf{294.9} & 395.4 & \textbf{102.7}\\
Sidecar scoring batches & --- & 350 & --- & 139\\
Repeated guidance deliveries & --- & 69 & --- & 4\\
\bottomrule
\end{tabular}

\caption{Frozen-controller transfer: task outcomes and score-normalized resource use. Cost rows divide the fixed 54-attempt workload's resource use by its paired mean benchmark score ($I_C$). M and K denote millions and thousands of tokens; dashes indicate no guidance mechanism.}
\label{tab:actor-transfer}
\end{table}

\paragraph{WebShop task gains and score-normalized efficiency.}
On the 53 jointly scored WebShop pairs, Synapse-Llama completes \textbf{three tasks versus none for Base-Llama} and raises the mean benchmark score from \textbf{3.09 to 14.34 (+11.25 points)}.
Input tokens per mean-score point decrease by \textbf{76.5\%}, from \textbf{4.94M to 1.16M}, and task time per mean-score point decreases by \textbf{74.0\%}, from \textbf{395.4 to 102.7 seconds}.
The request index decreases by \textbf{80.1\%}, and output tokens per mean-score point by \textbf{76.0\%}.
The frozen controller supports higher task yield and lower score-normalized resource use under a different model family without additional distillation.

\paragraph{AppWorld execution under the fixed action budget.}
AppWorld places greater demands on the actor's ability to complete and terminate multi-API workflows.
Neither arm fully solves a task among its 54 registered attempts.
The runs end at the 60-action cap or with an execution error, with no natural termination.
This pattern is consistent with limited workflow-completion capability in the evaluated Llama configuration: local contract guidance does not replace the actor's multi-step plan or termination decision.
On the 47 jointly scored pairs, both mean scores are \textbf{16.18}.
Synapse-Llama reduces input tokens per mean-score point by \textbf{4.2\%} (\textbf{2.30M versus 2.40M}), output tokens by \textbf{20.8\%} (\textbf{37.9K versus 47.9K}), and task time by \textbf{10.0\%} (\textbf{294.9 versus 327.8 seconds}) under the same normalization.
The observed benefit is lower score-normalized token use and task time at the same reported paired score.

\end{document}